\documentclass[a4paper,fleqn]{cas-sc}
\usepackage[authoryear]{natbib}
\usepackage[figuresright]{rotating}
\usepackage{graphicx}
\usepackage{subfigure}
\usepackage{caption}
\usepackage{tabularx}
\usepackage{float}
\usepackage[capitalise]{cleveref} 
\usepackage{subcaption} 
\usepackage{forest} 
\usepackage{xcolor} 
\usepackage{colortbl}  
\usepackage{placeins}
\definecolor{mypink2}{RGB}{219, 48, 122}
\definecolor{ForestGreen}{RGB}{34,139,34}
\newcommand{\citecustom}[1]{(see \citeauthor{#1}, \citeyear{#1})}

\makeatletter
\def\ps@pprintTitle{%
 \let\@oddhead\@empty
 \let\@evenhead\@empty
 \def\@oddfoot{\footnotesize\itshape
       {Preprint submitted to \ifx\@journal\@empty Elsevier
       \else\@journal\fi}\hfill\today}%
 \let\@evenfoot\@oddfoot}
\makeatother

\begin{document}
\let\WriteBookmarks\relax
\def\floatpagepagefraction{1}
\def\textpagefraction{.001}
\let\printorcid\relax   

\shorttitle{Learning for routing}

\shortauthors{Zhou et~al.}

\title [mode = title]{Learning for routing: A guided review of recent developments and future directions}

\author{Fangting Zhou\textsuperscript{\it a}\textsuperscript{,*}}

\author[2]{Attila Lischka}

\author[2]{Balázs Kulcsár}

\author[1]{Jiaming Wu}

\author[3]{Morteza {Haghir Chehreghani}}

\author[4, 5]{Gilbert Laporte}


\affiliation[1]{organization={Architecture and Civil Engineering, Chalmers University of Technology},
    city={Gothenburg},
    country={Sweden}}

\affiliation[2]{organization={Electrical Engineering, Chalmers University of Technology},
    city={Gothenburg},
    country={Sweden}}

\affiliation[3]{organization={Computer Science and Engineering,  Chalmers University of Technology and University of Gothenburg},
    city={Gothenburg},
    country={Sweden}}

\affiliation[4]{organization={Decision Sciences, HEC Montréal},
    city={Montréal},
    country={Canada}}

\affiliation[5]{organization={Management, University of Bath},
    city={Bath},
    country={United Kingdom}}


\begin{abstract}
This paper reviews the current progress in applying machine learning (ML) tools to solve NP-hard combinatorial optimization problems, with a focus on routing problems such as the traveling salesman problem (TSP) and the vehicle routing problem (VRP). Due to the inherent complexity of these problems, exact algorithms often require excessive computational time to find optimal solutions, while heuristics can only provide approximate solutions without guaranteeing optimality. With the recent success of machine learning models, there is a growing trend in proposing and implementing diverse ML techniques to enhance the resolution of these challenging routing problems. We propose a taxonomy categorizing ML-based routing methods into construction-based and improvement-based approaches, highlighting their applicability to various problem characteristics. This review aims to integrate traditional OR methods with state-of-the-art ML techniques, providing a structured framework to guide future research and address emerging VRP variants.
\end{abstract}


\maketitle

\renewcommand{\thefootnote}{}%
\footnotetext{* Corresponding author. Email: fangting@chalmers.se}%
\renewcommand{\thefootnote}{\arabic{footnote}}  

\section{Introduction}

In recent years, the intersection of Operations Research (OR) and Computer Science (CS) has garnered increasing attention. OR, with its focus on mathematical modeling and optimization, has long been instrumental in addressing transport challenges across various domains. Meanwhile, CS, particularly with the advent of machine learning (ML), has revolutionized approaches to problem-solving by learning from data and adapting to dynamic environments. Despite the potential synergy between OR and CS, there remains a noticeable gap in the integration of methodologies and techniques from these fields. Traditional OR methods, while effective in many scenarios, often struggle to cope with the scale and complexity of hard problems. Conversely, while ML techniques offer promising solutions, their application to optimization has been relatively underexplored within the OR community. In this paper, we concentrate on the application of ML to two classical and difficult problems in the field of OR: the Traveling Salesman Problem (TSP) and the Vehicle Routing Problem (VRP).

The goal of the TSP is to find a tour, given a set of cities $\{1, \dots, n\}$, that visits each city exactly once, ends where it started, and minimizes the total traveled distance. Typically, the problem assumes direct travel is possible between any pair of cities. The problem can be represented as a complete graph $G = (V, E)$, where $V = \{1, \dots, n\}$ and $E = \{ \{u, v\} | u, v \in V, u \neq v \}$. The objective is to find the cheapest Hamiltonian cycle. In our setting, we use Euclidean distance as the cost metric, and the travel costs are symmetric. However, there are variants of the TSP with asymmetric costs. Basic variants of the TSP include time windows, prize-collecting, or orienteering. The time windows variant (TSPTW) requires each city to be visited within a specified time frame. The prize-collecting variant (PC-TSP) assigns a reward or cost to each city, aiming to maximize total rewards while minimizing travel costs. The orienteering variant (OPT) combines elements of the prize-collecting TSP with a fixed time or distance limit, challenging the traveler to collect as many rewards as possible within these constraints.
While the origins of the TSP are somewhat obscure \citecustom{lawler1985}, its first treatment by means of OR methods clearly dates back to the study of \cite{dantzig1954} which proposes the first mathematical programming formulation and solution methodology for the problem. This has been followed by numerous studies based mostly on the application of branch-and-cut methods, culminating with the algorithm of \cite{applegate2006} which can optimally solve instances of several thousand cities. In parallel, several heuristics have been proposed, based mostly on $r$-opt moves which consist of successively removing $r$ cities from the tour, and reinserting them in different positions. The $r$-opt heuristic was formalized by \cite{lin1965} and extended by \cite{lin1973}. Recent computer implementations of the Lin-Kernighan heuristic \citep{helsgaun2000,taillard2019,taillard2022} can solve to near-optimality instances involving millions of cities.

The VRP is a generalization of the TSP and involves determining optimal routes for a fleet of vehicles to service a set of customers. There is a depot city (designated as city 0 for convenience) which can be visited multiple times. The other cities $\{1, \dots, n\}$ (interpreted as customers) each have a specific demand that must be fulfilled. These demands typically represent quantities of goods that need to be delivered. The objective of the VRP is to determine an optimal set of routes for a fleet of vehicles to deliver these goods, minimizing the total travel cost while ensuring that each customer's demand is met and that the routes start and end at the depot. A common version of the VRP is the Capacitated Vehicle Routing Problem (CVRP), which introduces an additional constraint: each vehicle has a limited carrying capacity and the vehicle fleet is homogeneous. Some common variants of the VRP include pick-up and delivery, time windows, or split deliveries. The pick-up and delivery (PDP) variant involves transporting goods not only from the depot to the customers but also picking up items from customers and delivering them either to other customers or back to the depot. The time windows (VRPTW) variant introduces specific time frames during which deliveries must occur, adding a temporal constraint that requires careful scheduling to ensure all deliveries are made within the designated windows. The split deliveries (SDVRP) variant relaxes the constraint that each customer can only be visited once, allowing a customer’s demand to be split across multiple deliveries. These variants introduce additional constraints and complexities, making the CVRP a rich and challenging problem in the field of transportation and logistics.
The VRP was introduced by \cite{dantzig1959} under the name "The Truck Dispatching Problem" and was later renamed "The Vehicle Routing Problem" by \cite{christofides1976}. The VRP is notoriously more difficult to solve than the TSP and has been mostly solved by means of heuristics based on various construction and improvement techniques. Nowadays, the best heuristics are hybridizations of a variety of metaheuristics that combine mathematical programming and large neighborhood search, decomposition, and cooperation techniques. These can routinely yield solutions whose values are close to those of large-scale benchmark instances \citep{laporte2014}. For recent implementations, the reader is referred to \cite{christiaens2020}, \cite{accorsi2021}, and \cite{santini2023}. There is now a tendency to develop heuristics that perform well on several VRP variants by using the same structure and the same parameters \citep{vidal2014}. The best exact algorithms are mostly based on branch-and-cut-and-price and can solve instances containing a few hundred cities \citep{poggi2014,pecin2014,pecin2017,pessoa2020}.

Several survey papers address ML for routing problems. Most reviews take a broader perspective, focusing on combinatorial optimization \citep{bengio2021,karimi2022,mazyavkina2021} and specific methodologies such as reinforcement learning \citep{mazyavkina2021}, deep reinforcement learning \citep{li2021research}, and graph neural networks (GNNs) \citep{cappart2023}. Other reviews narrow their focus to specific domains where routing is one of the key concerns, like manufacturing \citep{zhang2023review} and transportation \citep{farazi2021} or particular issues within the field, such as dynamic routing \citep{hildebrandt2023}. Some reviews specifically address routing problems, focusing on deep learning \citep{sui2025} or deep reinforcement learning \citep{wang2021deep}.

Four review papers specifically examine ML for routing \citep{li2022overview,bai2023,bogyrbayeva2024,zhou2024}, making them the most relevant to this topic. \cite{li2022overview} provide an overview and experimental analysis of VRP learning-based optimization algorithms, with a primary focus on reinforcement and supervised learning. \cite{bai2023} present a comprehensive review of hybrid methods that combine analytical techniques with ML tools in addressing VRP problems. \cite{bogyrbayeva2024} review ML methods for solving VRPs with a taxonomy that includes both end-to-end learning and hybrid models. \cite{zhou2024} propose a scientometric review of learning-based optimization (LBO) algorithms for routing problems. Nevertheless, the current paper stands out for its comprehensive and timely analysis, covering the most recent studies in machine learning for routing problems from 2016 to 2025. Reviewing an unparalleled number of 253 articles (a detailed breakdown of the papers is presented in Table \ref{tab:journal_counts}), it provides an extensive overview of the latest trends and technological advancements. Employing a systematic review method ensures a rigorous evaluation, offering new insights and potential solutions tailored to the specific challenges of routing. This focus fills a significant gap in the existing literature and introduces innovative methodologies. The detailed and updated analysis presented makes this paper an invaluable resource for researchers and practitioners aiming to apply cutting-edge ML techniques to solve complex routing issues. A summary of these review papers on learning and routing is presented in Table \ref{tbl1}.

\begin{table}[width=\linewidth,cols=2,pos=h]
\caption{{Release channel and number of papers}\label{tab:journal_counts}}
\begin{tabular*}{\tblwidth}{@{} LL@{} }
\toprule
        \textbf{Open access platform} & \textbf{Number of preprints}\\
        arXiv & 29\\
        \hline
        \textbf{Conference} & \textbf{Number of papers}\\
        Advances in Neural Information Processing Systems & 14 \\
        Proceedings of the AAAI Conference on Artificial Intelligence & 6\\
        International Conference on Learning Representations & 8\\
        \hline
        \textbf{Journal} & \textbf{Number of papers}\\
        European Journal of Operational Research & 13 \\
        Transportation Research Part C: Emerging Technologies & 11 \\
        IEEE Transactions on Intelligent Transportation Systems & 9 \\
        Transportation Research Part E: Logistics and Transportation Review & 8 \\
        Computers \& Operations Research & 7 \\
        Transportation Science & 7 \\
        Computers \& Industrial Engineering & 4 \\
        Expert Systems with Applications & 4\\
        Operations Research & 3\\
        OR Spectrum & 3\\
        Ocean Engineering & 3 \\
        IEEE Transactions on Cybernetics & 3 \\
        IEEE Transactions on Neural Networks and Learning Systems & 3\\
        IEEE Transactions on Artificial Intelligence & 2 \\
        INFORMS Journal on Optimization & 2 \\
        INFORMS Journal on Computing & 2\\
        Information Sciences & 2 \\
        International Journal of Production Research & 2\\
        Transportation Research Part B: Methodological & 2\\
        Transportation Research Part D: Transport and Environment & 2\\
        Sustainable Cities and Society & 2\\
        Networks & 2 \\
        Energy & 2 \\
        Engineering Applications of Artificial Intelligence & 2\\
        IEEE Transactions on Industry Applications & 2\\
    \bottomrule
\end{tabular*}
\end{table}

\begin{table}[width=\linewidth,cols=5,pos=h]
\caption{Summary of existing review papers on learning and routing}\label{tbl1}
\begin{tabularx}{\textwidth}{%
  >{\hsize=0.9\hsize}X
  >{\hsize=0.85\hsize}X
  >{\hsize=1.05\hsize}X
  >{\hsize=0.8\hsize}X
  >{\hsize=1.4\hsize}X
  }
\toprule
Article & Review timeline & \# Reviewed materials  & Areas of study & Research focus\\ 
\hline
\cite{bengio2021} & To 2019 & 71 papers  & Combinatorial optimization & ML builds partially learned combinatorial optimization algorithms\\ 
\cite{karimi2022} & 2000 - early 2021 & 136 papers & Combinatorial optimization & The integration of ML into meta-heuristics\\ 
\cite{li2022overview} & To early 2022 & 159 papers & Vehicle routing problem & Experimental analysis of Learning-based optimization algorithms (RL \& SL)\\
\cite{bai2023} &  To early 2022 & 212 papers & Vehicle routing problem & Hybrid methods combining analytical techniques and ML tools\\ 
\cite{bogyrbayeva2024} & To early 2023 & 129 papers & Vehicle routing problems & Divide ML (RL \& SL) into end-to-end learning and hybrid models\\ 
\cite{zhou2024} & To early 2024 & 153 papers & Routing problems & Bibliometric analysis of learning-based optimization algorithms\\
Current paper & Technology: 2016-2025, Applications: 2021-2025 & 253 papers   & Routing problems & ML recent advances and outlook\\
\bottomrule
\end{tabularx}
\end{table}

The main objective of this review paper is to serve as a user-friendly resource, bridging the traditional realm of OR with the contemporary landscape of ML. Our aim is threefold: Firstly, to merge these two distinct research areas by developing a clear understanding that brings together traditional OR methods with state-of-the-art ML techniques. Secondly, to assist researchers from both fields in gaining insights about each other's work.
Finally, it guides future research endeavors by identifying gaps, challenges, and opportunities, thus providing a roadmap for further investigations. 

This paper addresses several critical needs in the current research landscape: diverse methods and frameworks, new application domains, powerful implementation tools, and the spread of research. Firstly, the explosion of ML frameworks and techniques has introduced a plethora of methods for tackling routing problems, ranging from supervised and unsupervised learning to reinforcement learning and neural network-based approaches. Understanding and flexibly applying these diverse methods is crucial for researchers and practitioners. Secondly, the scope of routing problems has expanded beyond traditional domains, with significant developments in areas such as multimodal transportation, which involves integrating different modes of transport. This expansion presents new challenges and opportunities that require novel solutions facilitated by ML. Thirdly, the rise of robust implementation tools has made it easier to apply and test ML algorithms in real-world routing scenarios. These tools enhance the practical applicability of theoretical advancements, making it essential to review and highlight their capabilities and limitations. Lastly, there has been a substantial increase in research papers on ML applications in routing, particularly within transport journals. Visualizing this growth through relative percentage distributions can provide insights into trends and the significance of this research area.

To systematically categorize these ML methodologies, we propose a comprehensive taxonomy for ML-based routing methods, facilitating a clearer understanding of the landscape and aiding in identifying research gaps. Our proposed classification divides ML-based routing approaches into two primary categories: construction-based approaches, which construct routes from scratch and include one-shot and incremental methods, and improvement-based approaches, which enhance existing routes through iterative improvements. Improvement-based approaches are further divided into exact-algorithm-based methods that aid classical algorithms, heuristic-based methods that optimize subparts of solutions, and subproblem-based methods that address specific subproblems within the larger routing context. In evaluating these studies, we considered factors such as the specific ML formulation used, the techniques applied, and the novelty of each approach. Additionally, we examine various characteristics of routing problems that influence the application and effectiveness of these ML approaches. These characteristics include vehicle heterogeneity, the use of electric vehicles, customer-specific time windows, and the need for dynamic and stochastic modeling to handle real-time changes and uncertainties. The integration of routing into broader problems, such as location-routing, further highlights the complexity and multidimensional nature of these challenges. By categorizing ML-based routing methods in this manner, we aim to provide a structured framework that highlights the diversity of approaches and the specific contexts in which they are most effective. This taxonomy serves as a foundational reference for researchers and practitioners, guiding future developments in the field of ML-driven routing optimization.

The main contributions of this paper are:
\begin{itemize}
    \setlength{\itemsep}{0pt}
    \setlength{\parsep}{0pt}
    \setlength{\parskip}{0pt}
     \item This paper uniquely integrates traditional OR methods with cutting-edge ML techniques, creating a comprehensive framework that bridges two distinct research areas.
    \item This survey comprehensively summarizes the latest machine learning approaches for solving various VRPs. 
    \item We propose a structured taxonomy for ML-based routing methods, categorizing them into construction-based and improvement-based approaches.
    \item We identify gaps, challenges, and opportunities in learning for routing, providing a roadmap for future ML-driven routing research.
\end{itemize}

The remainder of this paper is organized as follows. Section \ref{section2} provides the introduction and background on machine learning methods, setting the stage for their application in routing problems. Section \ref{section3} presents a taxonomy of the various learning methods used in routing. Section \ref{section4} focuses on specific learning techniques applied to basic routing problems. Section \ref{section5} explores learning methods in more practical, applied routing scenarios. Section \ref{section6} provides a detailed discussion of the performance of these methods, highlighting challenges and generalization issues. Section \ref{section7} outlines the proposed research agenda, identifying future directions for advancement in this field. Finally, Section \ref{section8} concludes the paper with a summary of key findings.

\section{Machine learning background} \label{section2}

The purpose of this section is to give a small introduction to machine learning (ML), i.e., to models (e.g., neural network types) and learning formulations commonly used in papers that tackle routing problems with the help of ML.

\subsection{Neural network architectures}

In the following, we present several neural network architectures. We focus on the most important ones that are frequently used in studies within the scope of our survey. Further, we give a small comparison of the advantages and disadvantages of the different architectures.

\subsubsection{Multilayer perceptrons (MLPs)}

MLPs are the simplest architecture in the realm of neural networks (NNs).
As the name suggests, MLPs consist of several layers. Each layer contains trainable weights or neurons. These neurons are organized in a matrix and can be multiplied by an input vector. By this, a new vector is obtained, which is put through a non-linear activation function. These activation functions (popular choices are ReLU, sigmoid, or tanh) are necessary for the network to be able to learn non-linear relationships.
Afterward, the output vector is passed on to the next layer, and the process is repeated. After several layers, the final output is contained, whose dimensionality depends on the task that shall be solved. For a \textit{d}-dimensional classification task, the output might be of dimension \textit{d}, and the biggest entry in the output vector is the class of the input. In a regression task, the output might be of dimension 1. MLPs (as any NNs introduced in this section) can be trained by the error backpropagation method, where the learnable weights of the network are updated according to their gradient with respect to a loss function (also called error function).

\subsubsection{Graph neural networks (GNNs)}

GNNs are a type of neural network designed to operate on graph-structured input data. This is realized by allowing information between layers to flow only along the input graph's edges.
GNNs compute feature vectors for the nodes in the input graph. The internal updates between layers happen by aggregating the feature vectors of neighboring nodes from the previous layer in a weighted manner, multiplying them by learnable weights and passing them through a non-linearity before combining this information with the information of a node's feature vector from the previous layer.
The way the information of neighboring nodes is weighted before combining it with a node's previous feature vector is dependent on the particular GNN version. In graph attention networks (GATs; \cite{velickovic2017graph, brody2021attentive}) the weight is computed dynamically with the help of the attention mechanism. In graph convolutional networks (GCNs; \cite{kipf2016semi}) the weight depends on the node degrees (giving higher emphasis on nodes with fewer neighbors). 
GNNs have been successfully used in many graph-related tasks such as node or graph classification (\cite{wu2020comprehensive}) and, as routing problems can often be interpreted as graph problems, have been used in many papers tackling routing with ML.

\subsubsection{Recurrent neural networks (RNNs)}

RNNs are a class of neural architectures designed to process sequential data. They do this by featuring ``hidden states'' which they update based on their inputs and remember when processing the next inputs. These hidden states can be interpreted as connections within the network over time. 
RNNs have been used in the routing context when processing ``tours'' or ``solutions'' to routing problems with neural architectures. 
A popular variant of RNNs are long short-term memories (LSTMs; \cite{lstm}) which were designed to tackle the problem of vanishing gradients over time.

\subsubsection{Transformers}

Transformers are a version of neural networks leveraging the attention mechanism (\cite{vaswani2017attention}).
Similar to RNNs, they can be used to process sequences of inputs, however, unlike RNNs that update an internal hidden state, transformers compute ``compatibility scores'' in the form of attention values that specify how much an input should focus on each other input when computing its output for the next layer. As a result, the ``order'' of the elements in the sequence does not matter (if no additional positional encodings are added). This makes transformers a good choice for computing, e.g., encodings for cities in a routing problem based on the cities' coordinates. Via GATs, transformers are related to GNNs (\cite{joshi2020transformers}).

\subsubsection{Comparison of neural architectures}

In the setting of our survey, MLPs are typically used to make ``small'' decisions within a pipeline or to project outputs of more powerful architectures to a final representation (e.g., mapping a graph representation generated by a GNN to a scalar value).
Another use case of MLPs is to serve as a function approximator in Q-learning (e.g., when the state space contains continuous values like in \cite{aljohani2021}).
RNNs are typically used to create neural representations of (sub)tours. These representations can then be fed to an MLP to predict the cost of the processed subtour, e.g., \cite{zong2022rbg}.
GNNs and transformers often serve in similar contexts: Generating a neural representation of a whole routing problem. Therefore, they can be interpreted as competing architectures.
Performance-wise, transformers are often more powerful, however, they also need large amounts of training data to maximize their performance (compare \cite{min2023unsupervised}).
A limitation of transformer models is their restriction to node-level inputs (such as city coordinates or customer demands), while GNNs can also process edge-level features.
This is, e.g., important in asymmetric routing settings where the distance between two cities can not be derived from Euclidean coordinates. 
As a result, GNNs serve as important architectures in such settings \citep{kwon2021matrix, lischka2024greatarchitectureedgebasedgraph}. 
We note that it is possible to combine different architectures for particular problems, e.g., by using GNNs as an initial method to generate node-representations from asymmetric problems and using transformers afterwards \citep{drakulic2024bq}.

\subsection{Learning formulations}
In the following, we present the three learning formulations (or learning paradigms) encountered in papers tackling routing problems. {After presenting the three different formulations, we provide a small comparison among them.

\subsubsection{Supervised learning (SL)}

In supervised learning, a dataset consisting of input data samples \textit{X} and corresponding target values \textit{Y} is given.
Data instances $(x_i, y_i)$ from the dataset are then passed through a neural model, e.g. neural networks, and the deviation of the model output $\hat{y}_i$ and the ground truth $y_i$ is used to compute an error metric.
This error is then used to update the model's internal parameters to better capture the relationship between data input and output.
In our routing context, SL might be used to predict edges that are part of the optimal tour in a routing problem instance. A drawback of SL is the need for labeled data, which can often be hard to obtain in reality. Routing problems are typically NP-hard problems, which means that target values for SL are computationally expensive to obtain.

\subsubsection{Unsupervised learning (USL)}

USL does not need any target labels for the learning process. Typical applications for USL are clustering algorithms where inputs are grouped according to their similarity, e.g., the k-means algorithm \citep{macqueen1967some}.
In the routing context, this can be important when grouping cities that are close to each other. Furthermore, some papers have managed to formulate loss functions that update a model's parameters to find promising edges in routing problems without the need for target labels, making them essentially unsupervised.

\subsubsection{Reinforcement learning (RL)}

In reinforcement learning, a learning agent moves through a state environment by choosing different actions. These actions trigger rewards or penalties, allowing the agent to learn which actions are good and which are bad, given the current environment state.
Agents can keep track of how good certain states, actions, or state-action pairs are by updating tables (compare Q-learning; \cite{watkins1989learning}) or, in case the state-action space becomes prohibitively large, updating function approximators. If the function approximators are neural networks, the term deep reinforcement learning (DRL) is used. An example of this is deep Q-learning (DQL; \cite{mnih2013playing}). Q-values (within tables or functions) indicate how good a certain action is given an environment state and the agent can then choose the best action given these values. Alternatively, instead of Q-values, it is possible to directly learn a policy that indicates which action to choose given an environment state using the REINFORCE algorithm \citep{williams1992simple}.
In our routing setting, the learning agent can move through the problem instance, and each action decides on the next city to visit. The state is the current partial tour of an already visited city and the reward or punishment is the (negative) traveled distance.

The description of abbreviations used in the ML context is provided in Table \ref{MLabbreviatons}.

\begin{table}[width=.6\linewidth,cols=6,pos=h]
\caption{Description of abbreviations used in machine learning context}\label{MLabbreviatons}
\begin{tabular*}{\tblwidth}{@{} LL@{} }
\toprule
Abbreviation & Description \\
\hline
ML & machine learning\\
SL & supervised learning\\
USL &  unsupervised learning \\
RL & reinforcement learning \\
DRL & deep reinforcement learning \\
NN & neural network \\
MLP & multilayer perceptron \\
GNN & graph neural network \\
GAT & graph attention network \\
GCN & graph convolutional network \\
RNN & recurrent neural network \\
LSTM & long short-term memory \\
LLM & large language model \\
\bottomrule
\end{tabular*}
\end{table}

\subsubsection{Comparison of learning formulations}

In the context of this paper, most considered works either use SL or RL.
USL methods are typically only used in subroutines that, e.g., try to cluster cities and by this create a divide-and-conquer-based approach. 
Utilizing fully USL-based formulations that directly solve the whole routing problem is generally hard due to the inherent complexity of the setting, such formulations usually lead to NP-hard optimization problems.
Learning pipelines based on SL can often be trained quickly, as the model is told exactly how its output should look. However, obtaining labeled data in routing contexts is often difficult or even computationally unfeasible.
In contrast, RL has the advantage of not requiring such labeled data in advance. This comes at the cost of needing an agent to explore an environment that represents the problem setting, figuring out by itself which decisions are good and which ones are not. Therefore, RL forms a contrast to SL where the model is told directly how to behave. As a result, RL can be more time and resource-consuming during training.
As a rule of thumb, SL seems to be a good option for rather small or ``simple'' routing problems with few side constraints such as TSP (which is still NP hard!).
RL, on the other hand, is the option of choice for more realistic settings that incorporate many side constraints and do not have customized, fast solvers (such as Concorde).

\subsection{Large language models}

One further topic that has gained enormous popularity in recent years is large language models (LLMs). 
Even though our study is focused on routing problems, we briefly discuss LLMs and relate them to the ML context of our work, especially since there are first papers using such LLMs to tackle routing problems.
For instance, \cite{liu2023can} innovatively proposed a hybrid approach that represents delivery routes as sentences and zones as words, enabling a Word2Vec-based model to learn drivers’ behavioral patterns. The learned zone sequence is then refined using intra-zone TSP, effectively bridging human experience and optimization.
In general, as their name indicates, LLMs are models with large amounts of parameters that have been trained on large amounts of data. 
While early language models were based on RNNs, nearly all state-of-the-art LLMs today are built on the transformer architecture \citep{vaswani2017attention}.
Even though the exact way how LLMs are trained can vary, and typically involves multiple steps (to fine-tune them for specific tasks), they are typically trained using self-supervised learning. This means that the data is split into input and label for the training.
In the context of LLMs, this can imply that a sentence found in the dataset is split into a prefix and a suffix. The model is then trained to predict the next words, given the prefix.
Due to LLMs being trained on vast amounts of data, they can often produce meaningful outputs for a variety of inputs and generalize to many unseen new inputs (although they often struggle with tasks requiring mathematical implications).
In the context of routing, this might mean that an LLM can be asked to create a meaningful tour between 5 cities that shall be visited during a road trip.
The LLM might know the real-world, pairwise distances between these cities and create a meaningful order to visit them, effectively solving a small-scale instance of TSP \citep{yang2024large, ye2024reevo, masoud2024exploringcombinatorialproblemsolving}.

\section{Overview of machine learning methods} \label{section3}

In this section, we will provide short tutorials on ML methods. The classification of the ML methods can be found in Fig. \ref{FIG:illustration}. In our taxonomy, routing methods are divided into construction-based approaches and improvement-based approaches, which can be either exact or approximated. Exact methods guarantee optimal solutions through algorithmic techniques that strategically explore feasible solutions. We include exact methods with ML in Section \ref{Exactsection}, as it is an emerging trend to integrate machine learning with traditional exact methods. This integration often enhances the performance of exact methods by leveraging the predictive capabilities of ML, marking an exciting development in the field. On the other hand, approximated methods provide near-optimal solutions and are typically used when exact methods face computational challenges, particularly with larger, more complex problems. Moreover, within both the construction-based and improvement-based categories, ML methods can be classified as either full/pure ML or hybrid. Full or pure ML methods rely entirely on machine learning algorithms to solve the problem, while hybrid methods combine machine learning with other algorithms or models, where ML contributes partially to the solution process. The level of hybridization can vary depending on the extent of integration with traditional methods like MILPs or heuristics. Thus, our taxonomy reflects the diverse ways machine learning can be incorporated into both construction-based and improvement-based approaches, ranging from full ML implementations to hybrid models that combine ML with other optimization techniques.

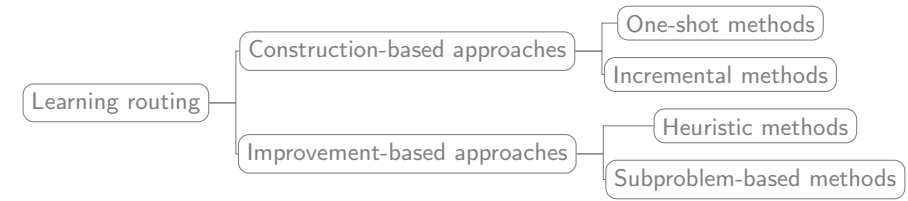
\begin{figure}[htbp]
    \centering
\begin{forest}
for tree={
    grow=east,
    draw, 
    rounded corners,
    node options={align=center},
    edge path={
        \noexpand\path [draw, \forestoption{edge}]
        (!u.parent anchor) -- +(10pt,0) |- (.child anchor)\forestoption{edge label};
    },
    parent anchor=east,
    child anchor=west,
}
[Learning routing
    [Improvement-based approaches
        [Subproblem-based methods]
        [Heuristic methods]
    ]
    [Construction-based approaches
        [Incremental methods]
        [One-shot methods]
    ]
]
]
\end{forest}
\caption{Proposed classification for ML methods for routing}
	\label{FIG:illustration}
\end{figure}

Apart from the ML methods included in Fig. \ref{FIG:illustration}, ML techniques can also be applied to algorithm selection and configuration for routing problems \citep{fellers2021,asinselecting}, which is beyond the scope of our work.
A further study that does not fall into any of our categories develops a metaheuristic \citep{baty2024}. 
The works of \cite{xin2021neurolkh} and \cite{zheng2021combining} do not directly fall into any of our categories since they use ML to improve existing heuristics (similar to how ML is used to improve methods exact-algorithm-based section).
However, due to the particular way ML has been incorporated in the pipelines of these papers, they can be included in the one-shot and heuristic methods subcategories.


\subsection{Construction-based approaches}

Construction-based approaches build solutions to the routing problem at hand from scratch.
 The category contains two subcategories: incremental methods and one-shot methods.

\subsubsection{Incremental methods}

In the subcategory of incremental methods, solutions are generated iteratively. 
This indicates that a solution is not built by applying a learned (sub)framework once, but multiple times. 
In practice, this means that, for example, a trained neural network is applied to a partial solution to output a new partial solution (which is closer to a complete solution than before) until it finally generates a complete solution.
Let's consider a small TSP instance of 5 cities \{1,2,3,4,5\}.
A partial solution could be the sequence of cities (4,3,5). 
This sequence can then be passed to the framework and we can receive a new sequence, where an additional, previously unvisited city was added, e.g., (4,3,5,2). We note that this new sequence is closer to a complete solution than before as it covers 4 out of 5 cities and only one additional city has to be added for the solution to be complete.
By passing the sequence (4, 3, 5, 2) to the framework once more, we receive the new output (4, 3, 5, 2, 1), which is a valid solution (although possibly non-optimal) to the given TSP.
We visualize this process for a graph with 20 nodes in \cref{fig:incremental}.
The blue path corresponds to the partial solution.
In the next iteration, a new city (green) is chosen among the remaining cities that need to be visited (other possible options are denoted by grey edges).

\begin{figure}[ht!]\centering
\begin{tabular}{ccc}
\includegraphics[width=0.31\textwidth, trim=4 4 4 4, clip]{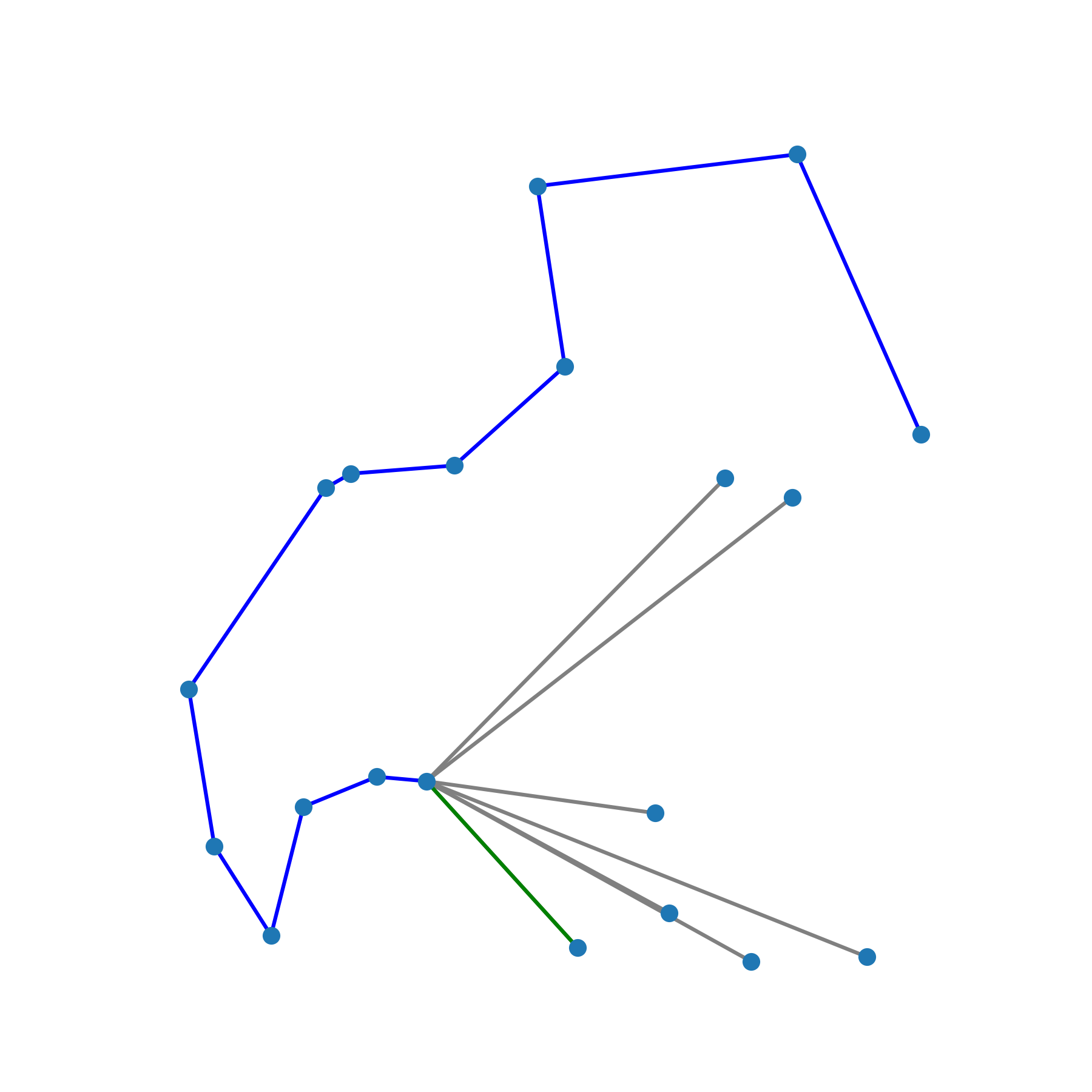} &
\includegraphics[width=0.31\textwidth, trim=4 4 4 4, clip]{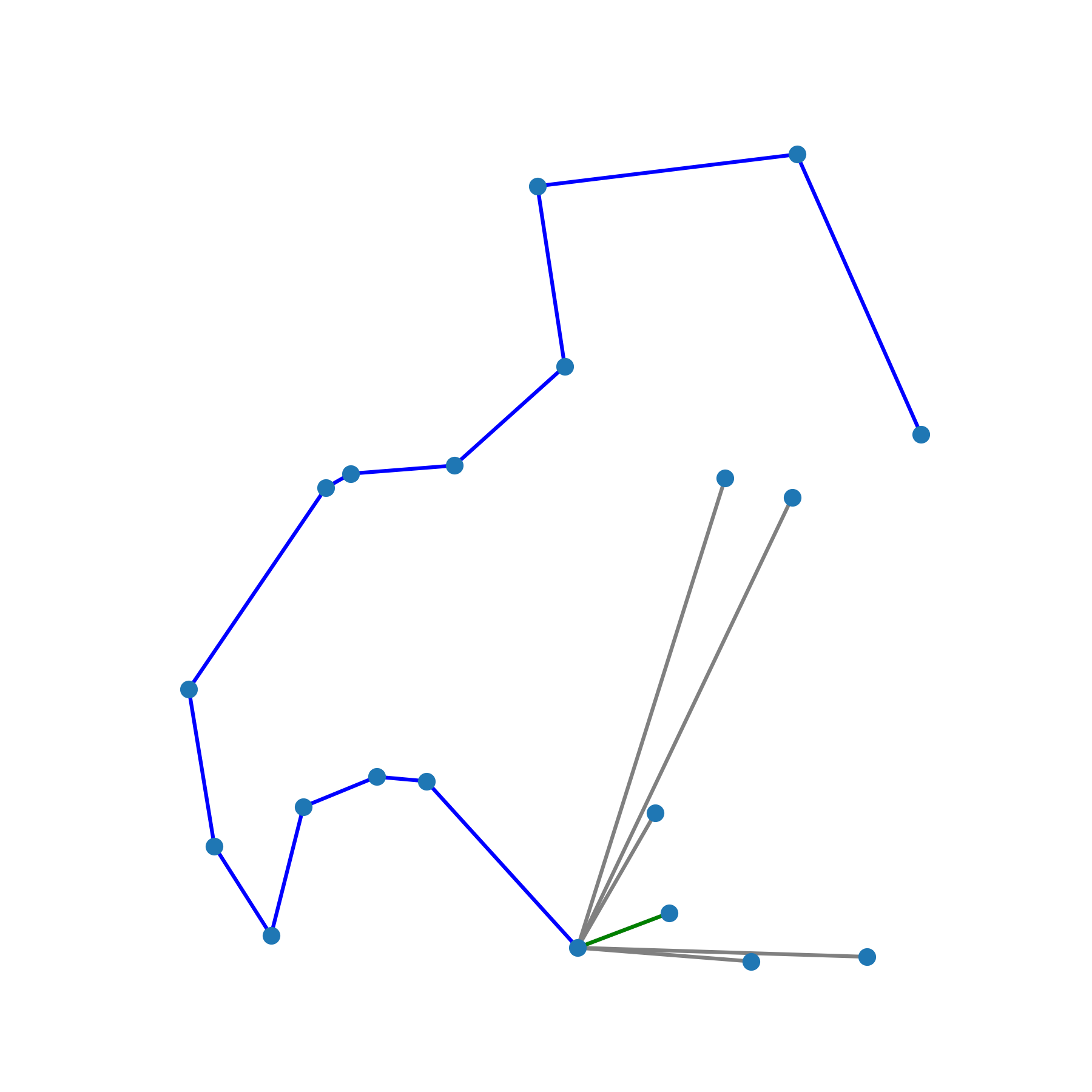} & \includegraphics[width=0.31\textwidth, trim=4 4 4 4, clip]{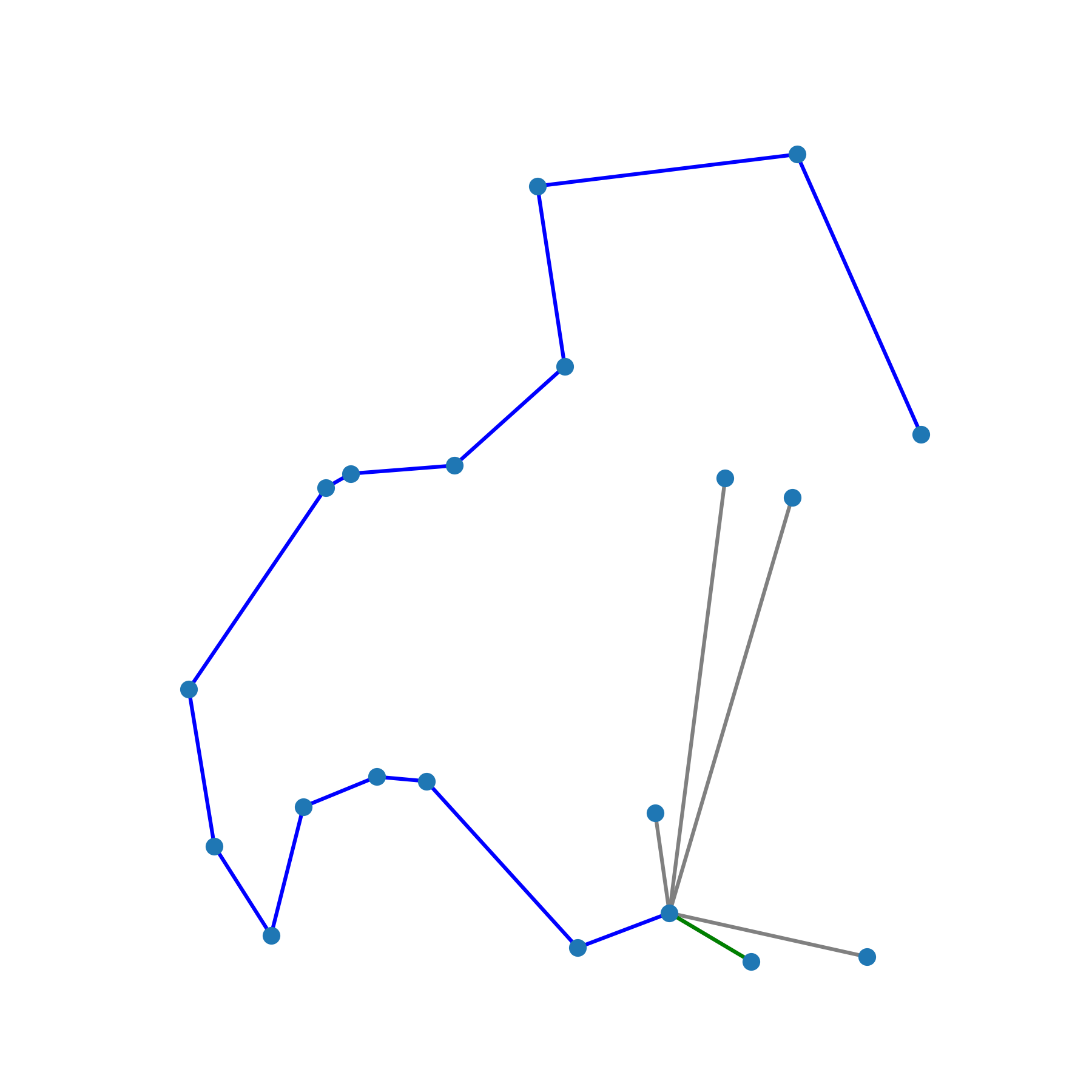} \\
(a) Step $n$ & (b) Step $n+1$ & (c) Step $n+2$
\end{tabular}
\caption{Incremental Methods - A trained model iteratively builds a solution.}
\label{fig:incremental}
\end{figure}

In practice, incremental approaches are typically two-fold, consisting of an encoder and a decoder framework.
The encoder computes feature vector representations for the given instance, e.g. embeddings for the different cities in the TSP instance. 
These embeddings typically attempt to capture the positions of the cities in a coordinate frame and the distances between them.
Given embeddings for the cities, it is possible to compute embeddings reflecting partial solutions, e.g., by averaging over the embeddings of the already visited cities in the partial solutions. 
A decoder framework can then, given an embedding of the partial solution, select the next city to visit.
With the new partial solution (where an additional city was added), a new partial solution embedding is computed and the decoder is applied again.
Depending on the complexity of the problem and its inherent constraints, different masking strategies might be necessary such that the partial solutions can still lead to an overall valid solution in the end.
For example, in the case of the TSP, it is not allowed to visit cities several times. In the case of the CVRP, it is possible to visit the depot multiple times, but it is necessary to ensure that the capacities of the vehicles are respected.

Another possibility for incremental approaches is learning and using lookup tables, e.g. Q-tables. Here, each possible partial solution (called state) has an entry in a table, indicating the next action to take (i.e., the next city to visit).
After adding the next city, the partial solution changes, and the lookup table is consulted again. 
Look-up tables can quickly grow prohibitively large, depending on how partial solutions are represented and how many possible actions remain.
In the case of routing problems, where exponentially many possible solutions exist, the lookup tables also grow exponentially in the size of instances.
Therefore, function approximators (e.g., Q-functions instead of Q-tables) are used, which can be similar to the before mentioned decoder architectures.

\subsubsection{One-shot methods}

In contrast to incremental methods, where learned frameworks are applied over and over until a solution is found, trained frameworks are only applied once in one-shot methods.
This does not mean, however, that a neural architecture directly outputs a valid solution to a given routing problem.
Instead, it typically means that the architecture generates an intermediate output which is afterwards used in a search algorithm.
The search algorithm is guided by the intermediate output and generates a valid solution.
A popular choice for intermediate outputs is probability heatmaps.
These heatmaps of size $n \times n$ for a TSP instance of size $n$, reflect the probability of moving from city $i$ to city $j$ at entry $(i,j)$ in the matrix.
For a TSP instance of 5 cities, a probability heatmap could look like this:
$$\begin{bmatrix}
0 & 0.98 & 0 & 0.02 & 0\\
0.2 & 0 & 0.76 & 0.04 & 0 \\
0 & 0 & 0.04 & 0. & 0.96 \\
1 & 0 & 0 & 0 & 0 \\
0 & 0 & 0.01 & 0.99 & 0 
\end{bmatrix}$$
This means the search algorithm would be nudged to find a solution that moves from city 1 to 2, from 2 to 3, from 3 to 5, from 5 to 4, and from 4 back to 1. 
Ideally, the matrix would only contain 1 and 0 values and reflect the optimal solution, making the additional search algorithm unnecessary.
Unfortunately, it is complicated to ensure that all problem constraints (guaranteeing) valid solutions by this, which is why additional search algorithms ensuring them are necessary. 
The intermediate outputs are generated by trainable neural frameworks that, similar to the encoders for the incremental approaches, try to capture the locations of cities in a coordinate system and the distances between them.
We visualize the process in \cref{fig:one_shot}.
In \cref{fig:one_shot} (a) we show the input TSP graph which is passed to the neural architecture. 
The output heatmap (in graph representation) can be observed in \cref{fig:one_shot} (b), where darker edges reflect higher probabilities.
In the end, we show the final solution found by a search algorithm in \cref{fig:one_shot} (c).

\begin{figure}[ht!]\centering
\begin{tabular}{ccc}
\includegraphics[width=0.31\textwidth, trim=4 4 4 4, clip]{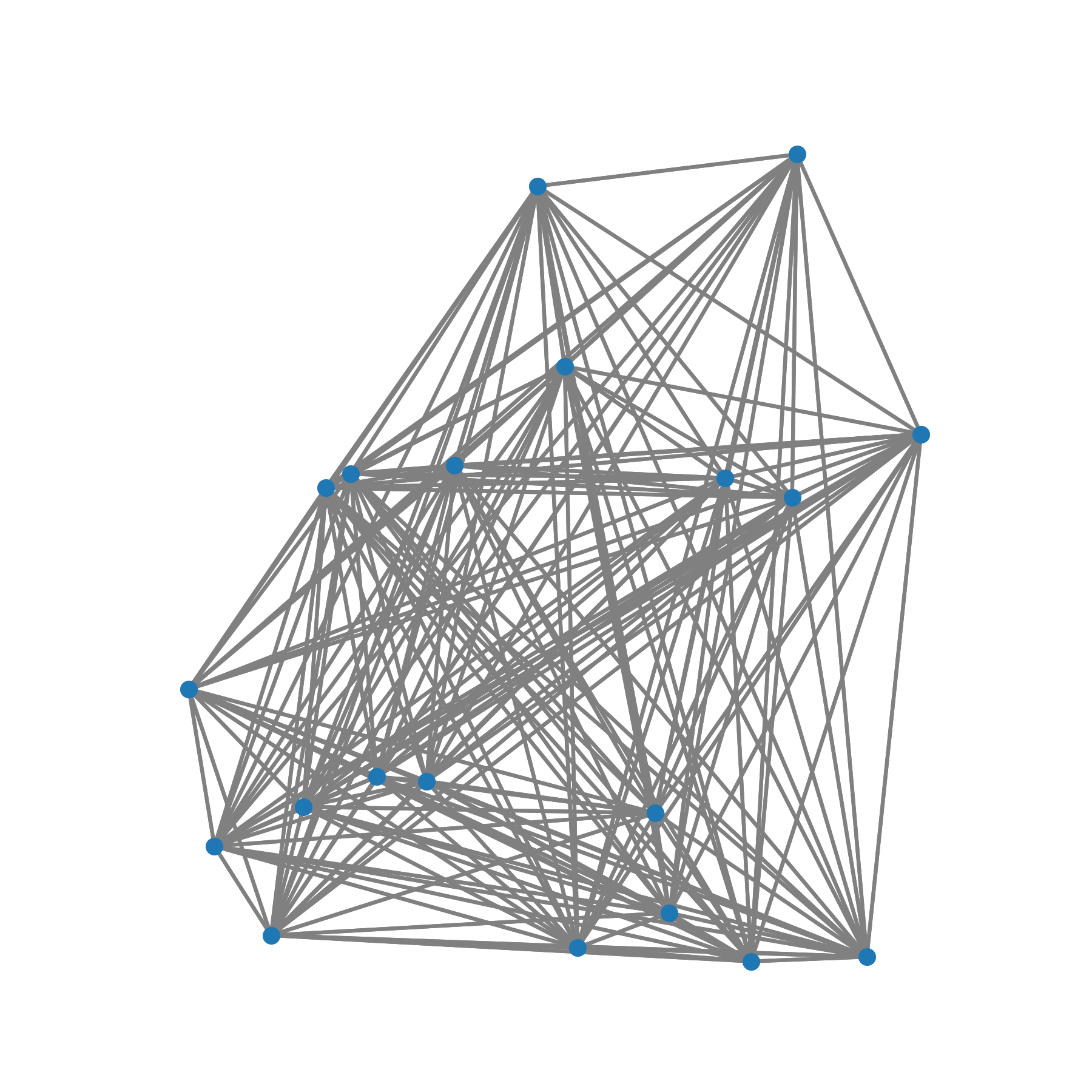} &
\includegraphics[width=0.31\textwidth, trim=4 4 4 4, clip]{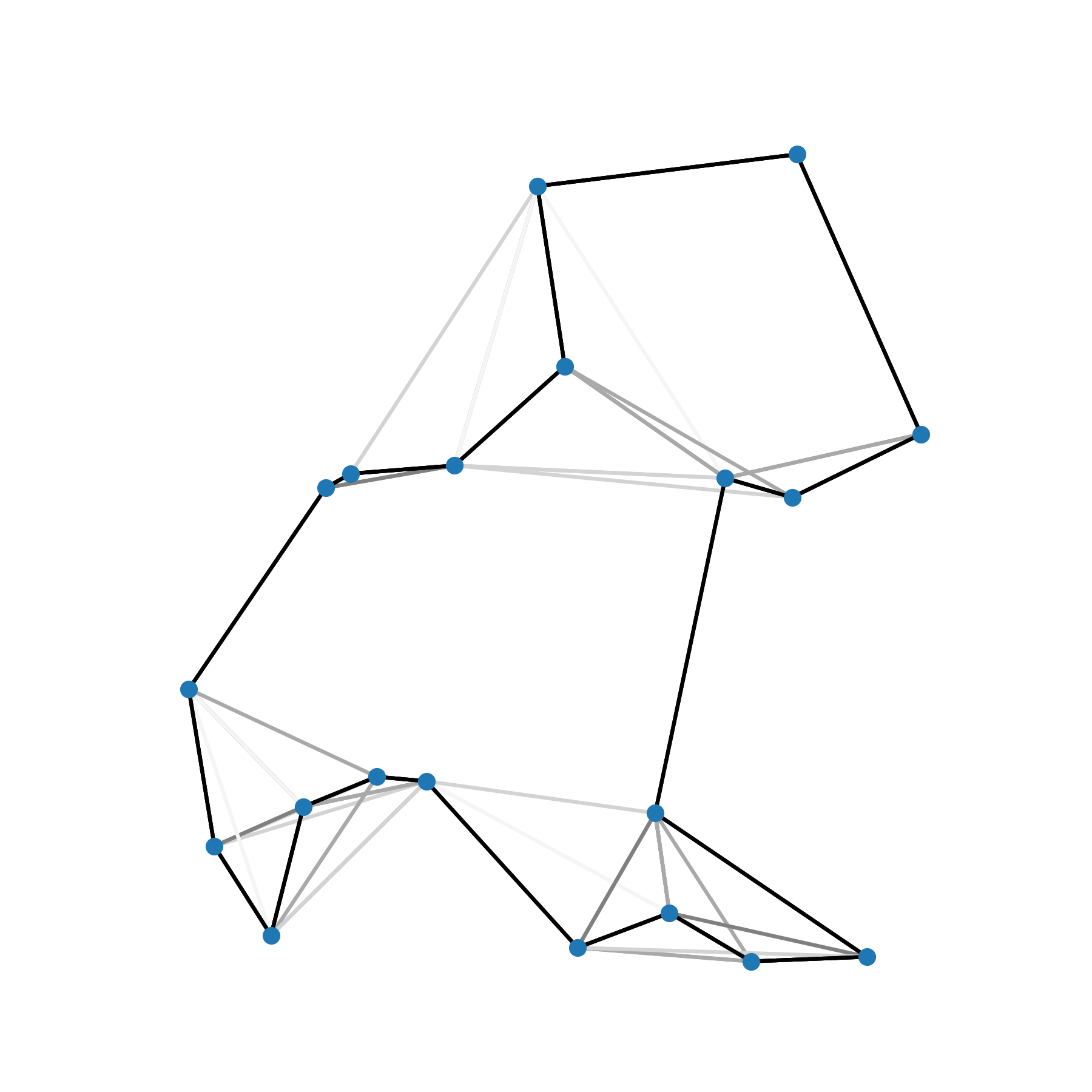} & \includegraphics[width=0.31\textwidth, trim=4 4 4 4, clip]{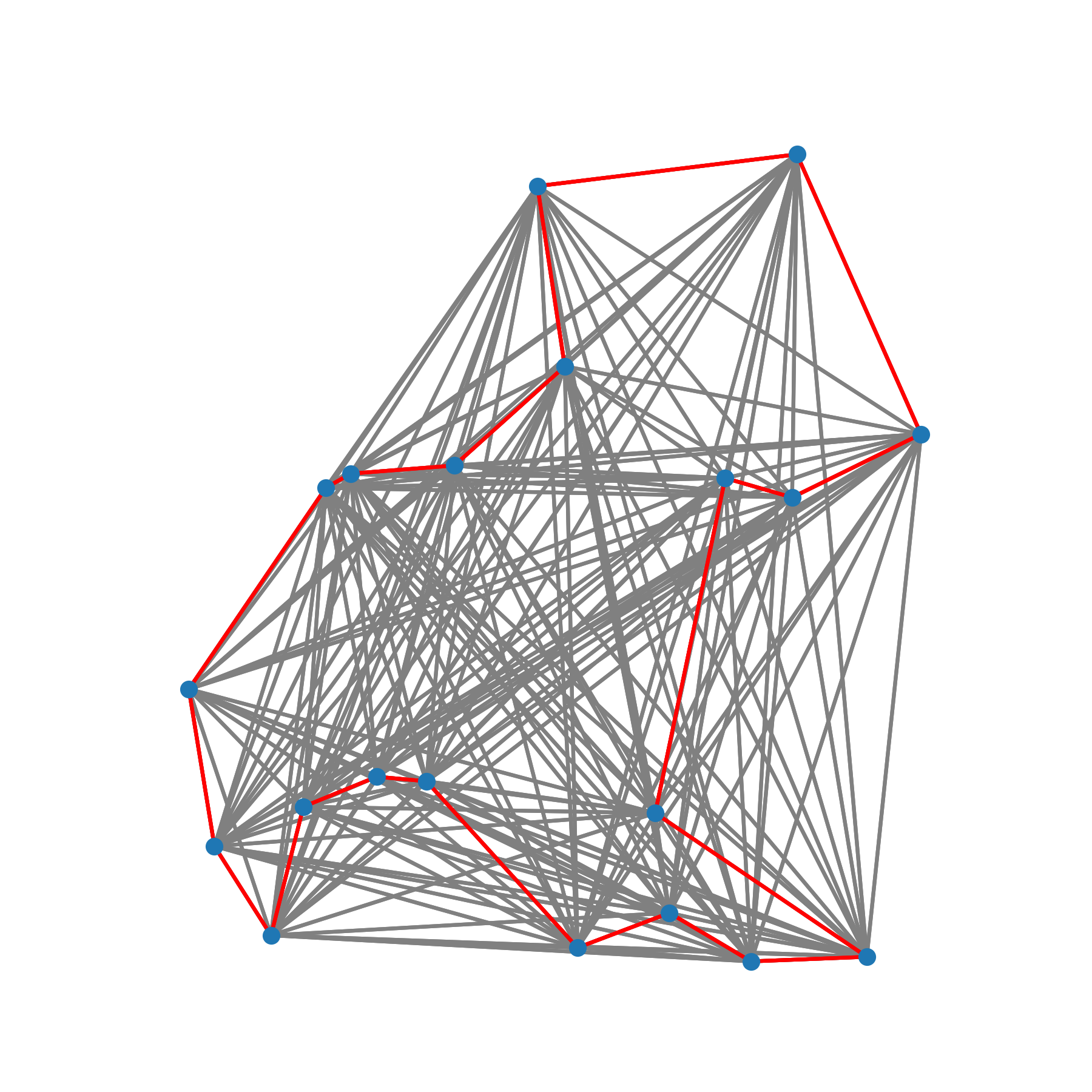} \\
(a) All Edges & (b) Heatmap & (c) Solution
\end{tabular}
\caption{One-Shot Methods - A trained model predicts an intermediate result, which helps find a solution.}
\label{fig:one_shot}
\end{figure}

\subsection{Improvement-based Approaches}

In contrast to construction-based approaches, where solutions are built from scratch, improvement-based approaches are built upon a given, valid initial solution, which is typically non-optimal and can be improved.
We split this category into three subcategories depending on how improvement operations are incorporated into the framework.

\subsubsection{Heuristic methods}

Heuristic methods are a subcategory of improvement-based approaches.
Here, heuristic refers to the way routing problems have been solved in the past to achieve (often non-optimal but good) solutions.
As we are in an improvement-based setting, we refer to improvement-based heuristics. 
An example of such a heuristic is $k$-opt. Here, $k$ edges in a current, valid solution (of a TSP, e.g.) are deleted and $k$ new edges are inserted while still guaranteeing the newly built solution's validity.
This is done in a way such that the new solution is better than the old one.
An example for $2$-opt can be found in \cref{fig:k-opt_heuristic}
This can be done iteratively until convergence, i.e., until no better solution is found.
In heuristic-based approaches in our learning to route setting, the rules according to which edges are deleted and substituted are learned by a neural architecture.
Such a neural architecture could, for example, consist of an encoder-like part capturing the problem instance and an additional ``output'' module that produces probabilities for the edges to be selected.
We note that $k$-opt is just a simple heuristic that can be incorporated in a learning framework but that there are many different possibilities to alter existing solutions (some of which can be expressed in, possibly several consecutive, $k$-opt moves).

\begin{figure}[ht!]\centering
\begin{tabular}{cccc}
\includegraphics[width=0.23\textwidth, trim=4 4 4 4, clip]{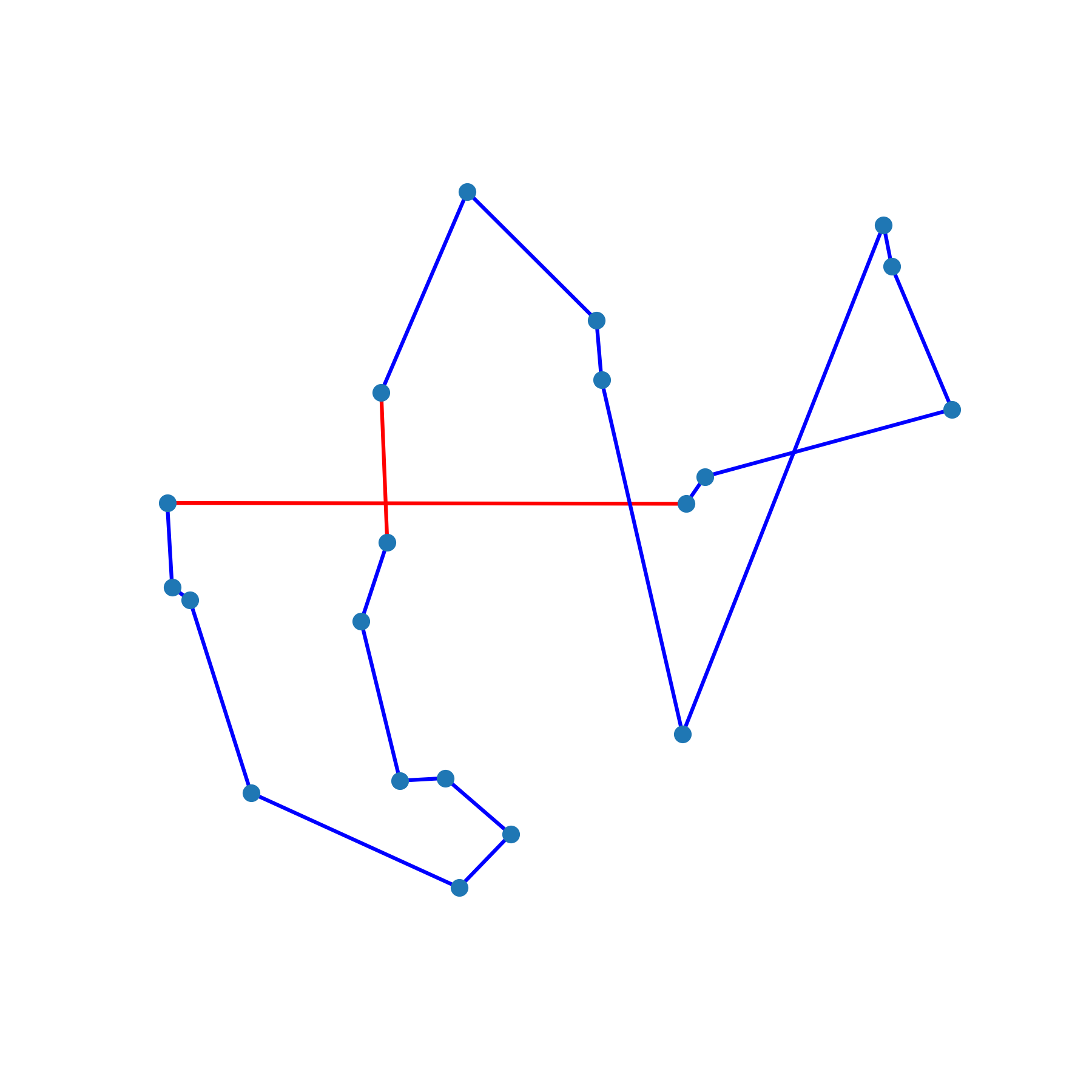} &
\includegraphics[width=0.23\textwidth, trim=4 4 4 4, clip]{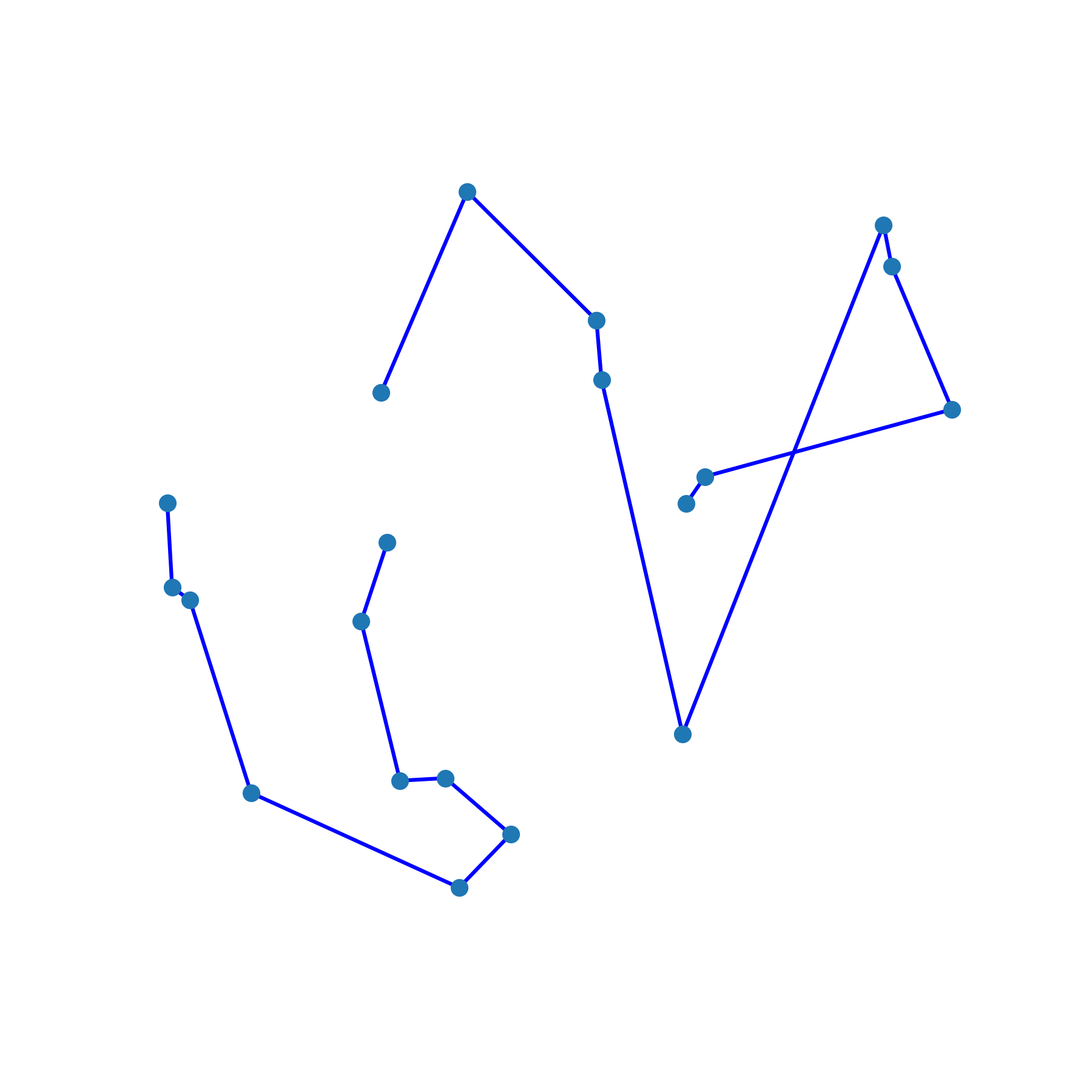} &
\includegraphics[width=0.23\textwidth, trim=4 4 4 4, clip]{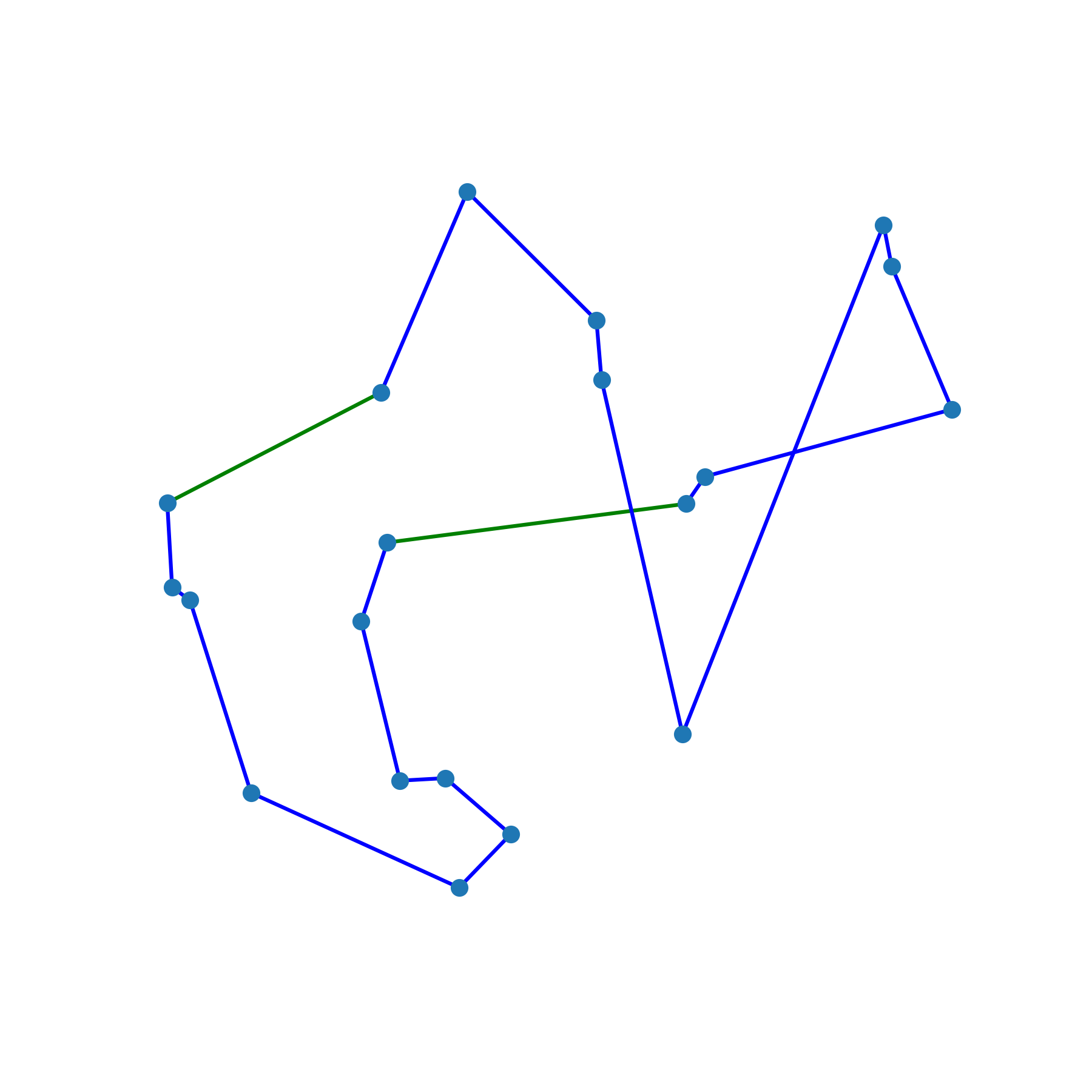} &
\includegraphics[width=0.23\textwidth, trim=4 4 4 4, clip]{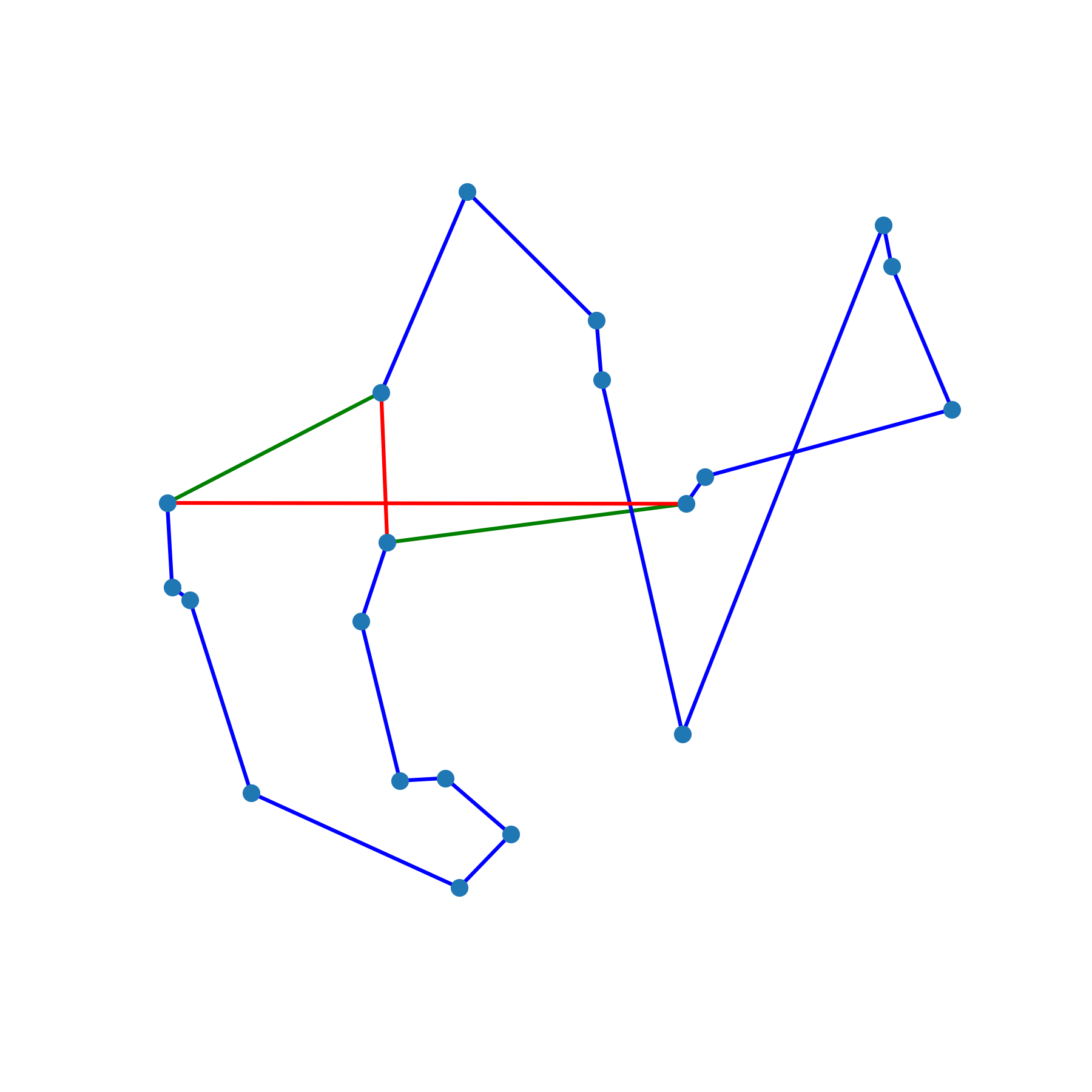} \\
(a) Select & (b) Delete & (c) Add & (d) Overview
\end{tabular}
\caption{$2$-opt visualized.}
\label{fig:k-opt_heuristic}
\end{figure}

\subsubsection{Subproblem-based methods}

In this improvement-based subcategory, the idea is to iteratively select subparts of a given current solution and optimize them.
The machine learning framework is either used to select the subpart that shall be optimized, to optimize a selected subpart or both.
This subcategory is especially suitable for large problem instances.
It can be applied easily to problems like the CVRP where several subtours exist that all start and end at a depot.
Selecting some of the subtours of a given solution and rearranging the cities in these subproblems to form another, valid subsolution and plugging them back into the overall problem, ensures that the overall problem solution also stays feasible while also respecting the overall solution's constraints, keeping a valid overall solution.
We visualize such a selection of subtours and optimize them in \cref{fig:subproblem}.
However, it is also possible to apply this idea to routing problems like the TSP. Here, a subsequence of a solution can be selected. 
This subsequence can then be optimized while ensuring that its start and end nodes stay the same. If the new subsequence has a lower cost than before, plugging it back into the overall solution also ensures that the overall cost decreases while still having a valid solution. 
An example can again be found in \cref{fig:subproblem}.

\begin{figure}[ht!]\centering
\begin{tabular}{cccc}
\includegraphics[width=0.23\textwidth, trim=4 4 4 4, clip]{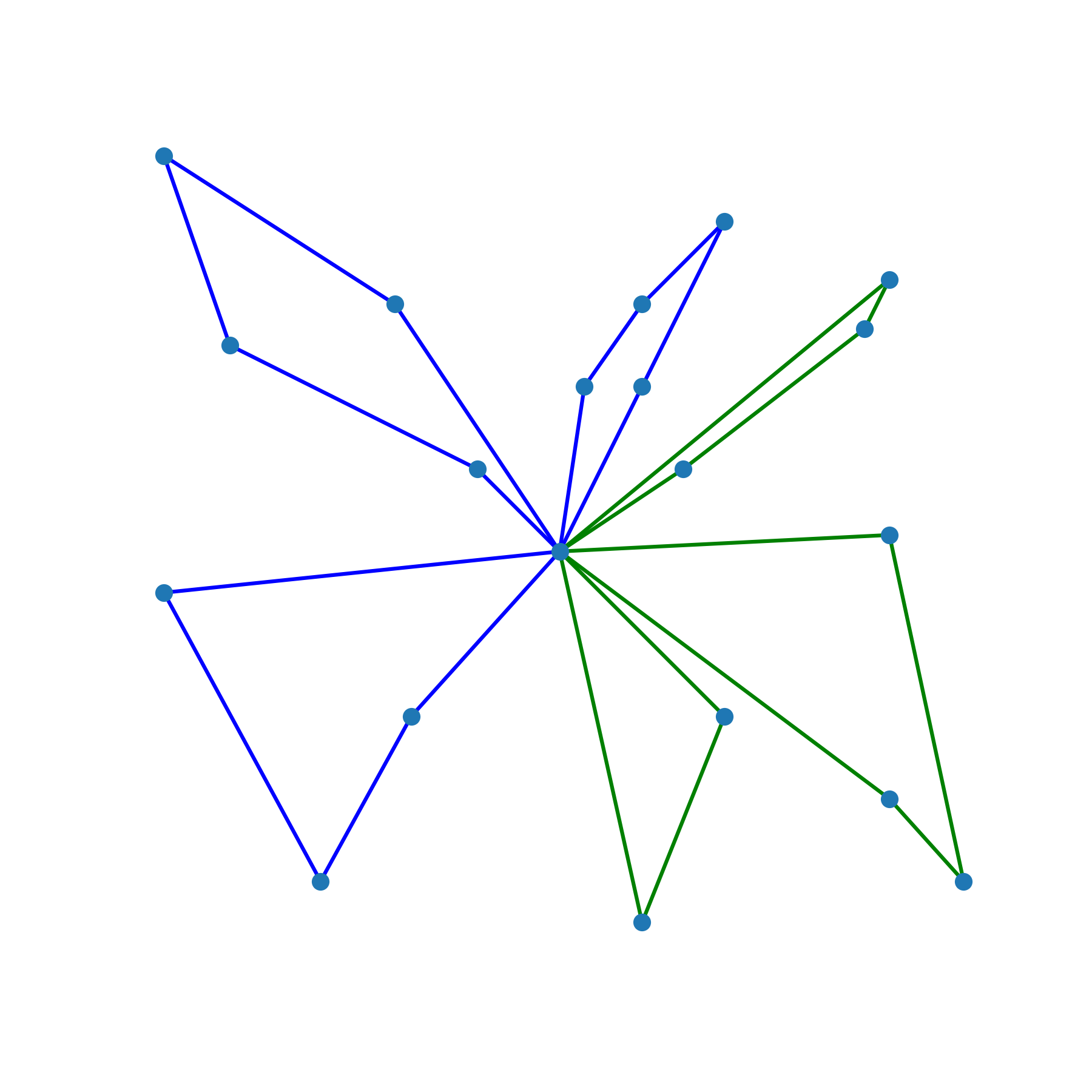} &
\includegraphics[width=0.23\textwidth, trim=4 4 4 4, clip]{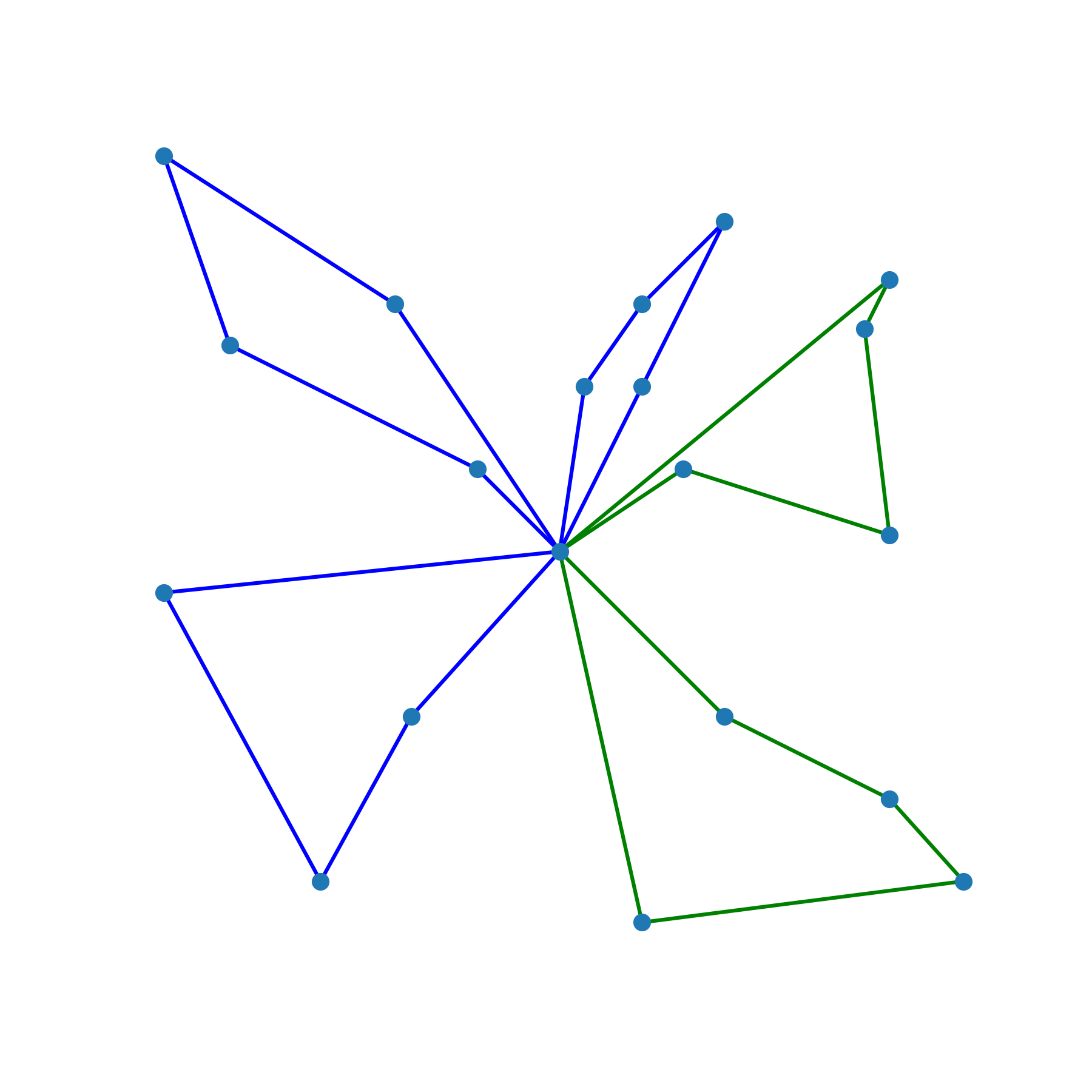} &
\includegraphics[width=0.23\textwidth, trim=4 4 4 4, clip]{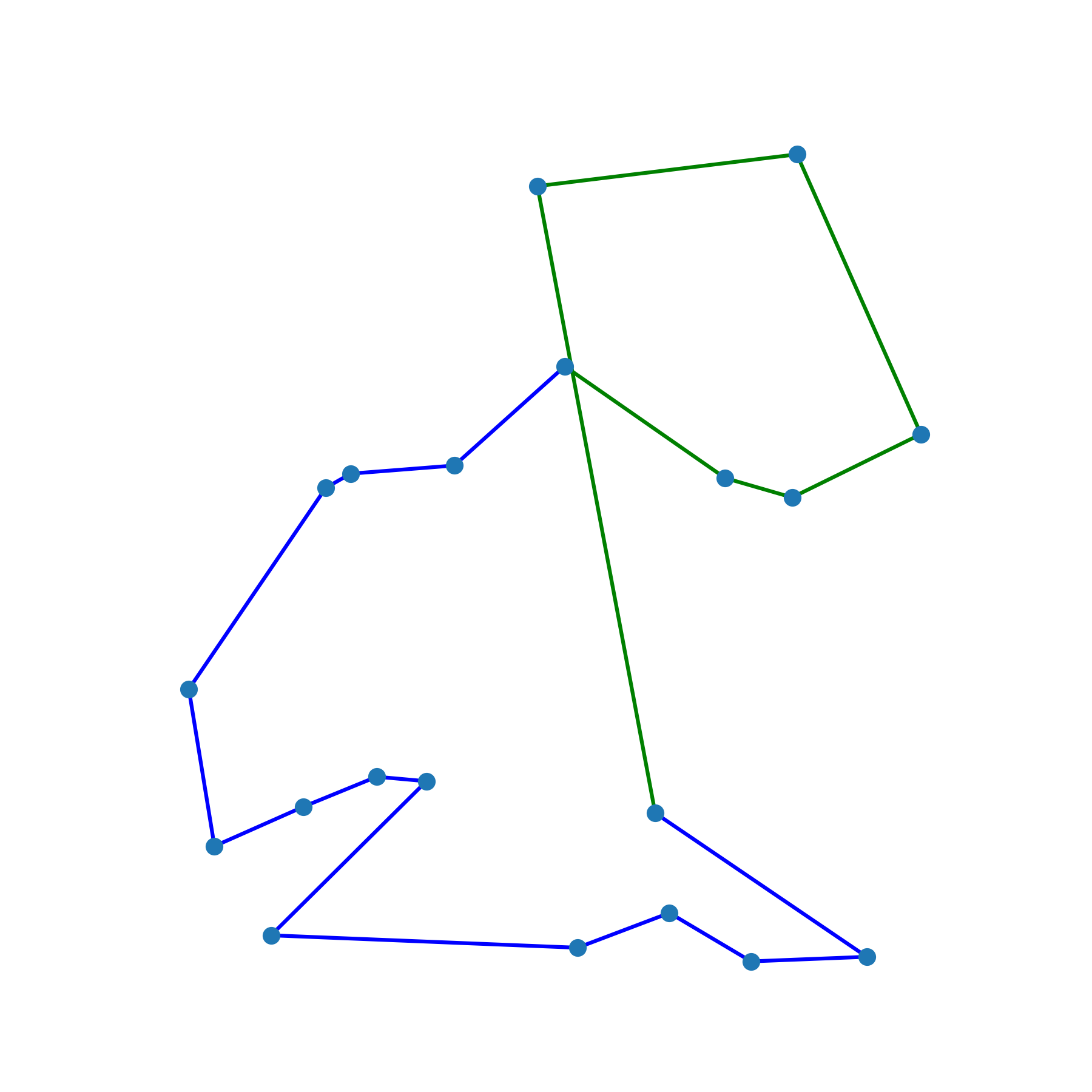} &
\includegraphics[width=0.23\textwidth, trim=4 4 4 4, clip]{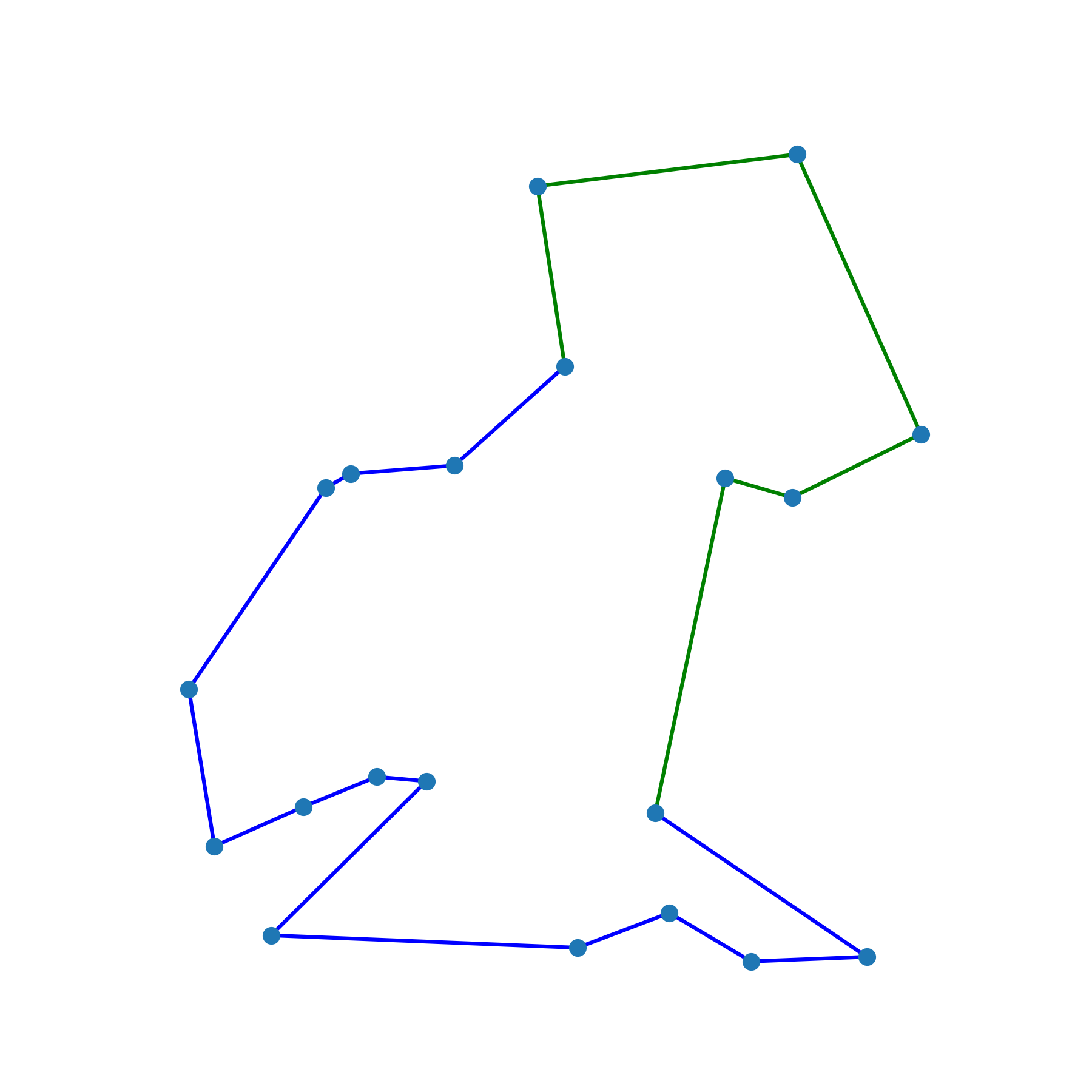} \\
(a) Selected VRP & (b) Optimized VRP & (c) Selected TSP & (d) Optimized TSP
\end{tabular}
\caption{Subproblem-Based Method.}
\label{fig:subproblem}
\end{figure}

\subsection{Exact-algorithm-based methods} \label{Exactsection}

Exact-algorithm-based methods refer to the integration of machine learning techniques with classical optimization algorithms to enhance their performance. These methods leverage ML to assist classical algorithms in solving routing problems by, for instance, reducing the search space, guiding the branching process of exact solvers, or predicting problem-specific parameters. Combining machine learning with exact methods can improve computational efficiency and scalability, especially when dealing with large, complex, or dynamically constrained problems.
Exact-algorithm-based methods can be interpreted as both, improvement and construction based. This is because we can interpret influencing, e.g., the branching process of an exact algorithm as a guidance to find new, better solutions. This corresponds to improving the best solution found for a problem instance so far. However, since we typically don't improve the old solution by modifying it but ``construct'' a completely new one, it can also be interpreted as construction-based or a hybridization of both approaches. 
Another way to integrate ML into exact-algorithm-based methods is to use a ML-based solver to find a good initial solution that gets passed to the optimal solver. Since this initial solution could be generated by both, construction-based and improvement-based approaches, exact-algorithm-based methods are again hybrid. Similarly, both incremental and improvement-based approaches can be used to build solutions for specific branches in the optimal solver, providing estimates on how to proceed in the optimal solution generation and making the approach hybrid.

\subsection{Comparison of machine learning methods}

In the following, we try to discuss the advantages and disadvantages of the different machine learning methods.
In general, it often holds that one-shot methods require expensive ground truth labels for training, which are hard to obtain. This is especially limiting for routing problems with additional constraints compared to basic TSP. 
During inference, one-shot methods are often slower, since additional search algorithms are required to decode the final solutions.
While it is possible to limit the runtime of such search algorithms, the final performance of the model might suffer. 
Incremental methods are highly popular since they can easily be trained using RL. Incrementally building the solution tours further allows us to consider more side constraints, as invalid moves can be masked out during solution creation.
Since solutions are created directly, additional search algorithms are optional. As a result, the trained frameworks can often create valid tours of high quality fast. 
A disadvantage is, however, that training incremental models often requires lots of resources in terms of GPU power and memory compared to single-shot models.
Heuristic and subproblem-based methods are best suited for very large instances, very specialized problems (possibly as a form of offline learning), or when desiring solutions of very high quality at the cost of potentially higher runtime.
This is because improvement-based approaches can be interpreted as a form of search algorithms that can be allocated large amounts of time for better performance.

\section{Machine learning methods in routing problems} \label{section4} 

In this section, we present various works related to the different categories of ML methods mentioned above. 
We focus on ``theoretical'' or ``methodological'' papers, dealing with basic routing problems such as the TSP and the CVRP without further assumptions usually encountered in real-life settings (e.g., time windows).
The papers that are presented here, focus on the technical part of creating frameworks to solve routing problems with the help of machine learning. 
The papers presented in this section helped form the different categories of Machine Learning Methods, which is why we group them again by these earlier introduced categories.
For all works that include publicly available code repositories, we provide links in our public code collection repository.\textsuperscript{1}
\renewcommand{\thefootnote}{}%
\footnotetext{\textsuperscript{1} https://github.com/Learning-for-routing/Repository-Collection}%
\renewcommand{\thefootnote}{\arabic{footnote3}}  

\subsection{Construction-based approaches}
Similar to the way the previous section was structured, we start with \textit{construction-based approaches} where solutions to routing problems are built from scratch. 

\subsubsection{Incremental methods}
We provide an overview of several studies proposing a diverse set of incremental methods in Table \ref{tableIncrementalCS} and Table \ref{tableIncrementalCS2}. Corresponding explanations for the abbreviations used in the tables can be found in Table \ref{abbreviatons}.

\begin{table}[width=\linewidth,cols=5,pos=ht]
\caption{Incremental methods in routing problems}\label{tableIncrementalCS}
\begin{tabularx}{\textwidth}{%
  >{\hsize=1\hsize}X
  >{\hsize=0.6\hsize}X
  >{\hsize=0.8\hsize}X
  >{\hsize=1.2\hsize}X
  >{\hsize=1.4\hsize}X
  >{\hsize=0.45\hsize}X
  }
\toprule
 & Problems & ML formulation & ML techniques  & Novelty & Code\\
\hline
\cite{bello2017neural} & TSP & RL (AC PG) & LSTM + PN & Overall framework & Unofficial \\
\cite{vinyals2017pointer} & TSP & SL & LSTM + PN & PN + overall framework & Unofficial \\
\cite{dai2018learning} & TSP & RL (DQL) & Structure2vec (GNN) & Overall framework & Yes \\
\cite{nazari2018reinforcement} & CVRP & RL (AC PG) & Emb enc., RNN + att. dec. & Overall framework & Yes \\
\cite{deudon2018learning} & TSP & RL (AC PG) & TF enc., PN dec. & Overall framework & Couldn't locate\\
\cite{kool2018attention} & TSP, CVRP & RL (PG, GBL) & TF enc., att dec. & Overall framework & Yes\\
\cite{kwon2020pomo} & TSP, CVRP & RL (PG, SBL) & TF enc., att dec.  & POMO & Yes\\
\cite{jin2023pointerformer} & TSP & RL (PG, SBL) & TF enc., multi-PN dec.& Generalization to large instances & Yes \\
\cite{lischka2024} & TSP & RL (PG, SBL) & TF enc., multi-PN dec. & Sparsification preprocessing & WIP\\
\cite{ling2023deep} & TSP & RL (PG, GBL) & CNN enc. and dec. & Overall (CNN-based) framework & Couldn't locate\\
\cite{zhang2020deep} & TSPTWR & RL  (PG, GBL) & TF enc, att dec. & Extension to TSP with TWs and rejection
& Couldn't locate\\
\cite{jiang2022learning} & TSP, CVRP & RL (PG, GBL), SL & TF enc., att dec., GCN enc. and dec. & DRO (generalization to non-uniform dists) & Yes \\
\cite{sultana2021learning} & TSP, CVRP & RL (PG, GBL), RL (AC PG) & TF enc., att dec., RNN + att. dec. & Improved learning by ER & Couldn't locate\\
\bottomrule
\end{tabularx}
\end{table}
\clearpage

\begin{table}[width=\linewidth,cols=5,pos=ht]
\caption{\textit{(Continued.)} Incremental methods in routing problems}\label{tableIncrementalCS2}
\begin{tabularx}{\textwidth}{%
  >{\hsize=1\hsize}X
  >{\hsize=0.6\hsize}X
  >{\hsize=0.8\hsize}X
  >{\hsize=1.2\hsize}X
  >{\hsize=1.4\hsize}X
  >{\hsize=0.45\hsize}X
  }
\toprule
 & Problems & ML formulation & ML techniques  & Novelty & Code\\
\hline
\cite{xin2020multidecoder} & TSP, CVRP & RL (PG, GBL) & TF enc., multiple att. dec. & Multi decoder and emb. glimpse & Yes\\
\cite{ma2019combinatorial} & TSP & RL (hierarchical PG, GBL &graph PN enc., att. dec. & Overall (hierarchical) framework & Yes \\
\cite{xing2020graph} & TSP & SL & GNN with PN & MCTS  & Yes\\
\cite{ouyang2021} & TSP & RL (PG, LS BL + CL) & GNN + MLP enc., att dec. & PG with LS and CL & Couldn't locate \\
\cite{sultana2022} & TSP & SL & CNN enc., LSTM + att dec. & Generalization to non-uniform dists. & Couldn't locate\\
\cite{xu2021} & TSP, CVRP & RL (PG, GBL) & modified TF enc., att dec. & Improved transformer encoders & Couldn't locate\\
\cite{bresson2021} & TSP & RL (PG, GBL) & TF enc., att dec. & New decoding context & Yes \\
\cite{mele2021reinforcement} & TSP & RL (AC PG) & TF enc., PN dec. & Incorporate MST info in loss & Couldn't locate\\
\cite{gunarathna2022} & dynamic TSP/CVRP & RL (PG, GBL) & Temporal/spatial TF-based & Extension to dynamic problems & Yes\\
\cite{hu2021} & TSP & SL & GNN + MLP enc. and dec. & Generalization to arbitrary symmetric graphs & Yes\\
\cite{zhou2023} & TSP, CVRP & RL (PG, SBL) & TF enc., att dec. &  meta-learning for fast adaptation & Yes\\
\cite{perera2023} & TSP & RL (PG, GBL) & GNN + LSTM enc., PN dec. & Focus on multi-objective TSP & Couldn't locate\\
\cite{lin2022pareto} & TSP, CVRP & RL (PG, SBL) & TF enc., modified att. dec. & Multi-objective preference for dec. & Yes\\
\cite{li2020deep} & TSP & RL (PG, AC) & 1DCNN enc, RNN + PN & MO, learn models for different subproblems & Yes \\
\cite{ruiz2023} & PCTSP & RL (Q-Learning) & Q-Table & Online Multi-Agent RL for PCTSP & Couldn't locate\\
\cite{mele2021} & TSP & SL and RL & CNN & Using ML in a construction heuristic & Yes\\
\cite{jiang2024} & TSP, CVRP & RL (PG, SBL) & TF enc, multiple att dec. & Several decoders for several dists. & Couldn't locate\\
\cite{pirnay2024selfimprovement} & TSP, CVRP & Self-SL & TF based & Train model with own sampled solutions  & Yes\\
\cite{yang2023memory} & TSP & RL (PG, GBL) & Custom TF & New TF model for large instances & Yes\\
\cite{kwon2021matrix} & TSP & RL (PG, SBL) & Matrix enc. + att. dec. & Matrix encoder, Asymmetric TSP & Yes \\
\cite{lischka2024greatarchitectureedgebasedgraph} & TSP & RL (PG, SBL) & GREAT enc. , multi-PN dec. & GREAT encoder, Asymmetric TSP & Yes \\
\cite{drakulic2024bq} & TSP, CVRP & SL & TF or TF + GNN & bisimulation quotienting in MDP & Yes 
\\
\cite{drakulic2024goalgeneralistcombinatorialoptimization} & TSP, CVRP & SL & custom TF & multi-task, generalist model & Yes \\
\cite{zhouMVMoE2024} & TSP, CVRP & RL (PG, SBL) & custom TF enc MoE, att dec. MoE & multi-task model for VRPs & Yes \\
\cite{rydin2025} & TSP, CVRP & RL (PG, SBL) & GREAT enc., modified att. dec. & multi-graph and MO setting on asymmetric VRPs  & Yes \\
\bottomrule
\end{tabularx}
\end{table}

\begin{table}[width=.6\linewidth,cols=6,pos=htbp]
\caption{Description of abbreviations used in tables summarizing routing papers}\label{abbreviatons}
\begin{tabular*}{\tblwidth}{@{} LL@{} }
\toprule
Characteristic & Description \\
\hline
AC & actor-critic\\
BL & Bandit Learning \\
BS & beam search\\
CL &  Curriculum Learning \\
DP & Dynamic Programming \\
DQL & deep Q learning \\
DRO & Distributionally Robust Optimization \\
ER & Entropy Regularisation \\
FW & framework \\
GBL & greedy baseline \\
GLS & Guided local search \\
GREAT & graph edge attention network \\
MCTS & Monte Carlo tree search \\
MO & multi-objective \\
MST & minimum spanning tree \\
LNS & large neighborhood search \\
LS BL & local search baseline \\
PG & policy gradient\\
PN & Pointer network \\
SBL & shared baseline \\
SR & subregion \\
SWA & stochastic weight averaging \\
TF & transformer \\
VNS  & Variable Neighborhood Search \\
MoE & Mixture-of-Experts \\
\hline
emb & embedding \\
att & attention \\
dists & distributions \\
enc & encoder \\
dec & decoder \\
samp & sampling \\
\bottomrule
\end{tabular*}
\end{table}

In an early work, \cite{vinyals2017pointer} introduce the pointer network to learn conditional probabilities of an output sequence like the sequence in an optimal TSP tour.
\cite{bello2017neural} is one of the first successful studies using a deep learning-based framework to tackle TSP.
\cite{dai2018learning} use structure2vec, a type of GNN, to learn a deep Q function for iteratively building solutions.
Another early work featuring a framework for incrementally building solutions to the CVRP using ML is \cite{nazari2018reinforcement}.
The authors map the input features of all customers (e.g. the demand) to a high-dimensional space.
These encodings are then passed to an LSTM and an attention mechanism is used to determine the next node that should be selected.
Their model is trained using RL.

Another early work by \cite{deudon2018learning} uses the attention mechanism (\cite{vaswani2017attention}) to compute encodings for the TSP.
Afterward, they use a pointing mechanism to select the next city on the tour. Similarly,
\cite{kool2018attention} also propose a transformer (attention) model which can also be interpreted as a graph attention network. 
They train their network to solve a variety of routing problems, including the CVRP and TSP. The work of \cite{kool2018attention} has since served as a basis upon which many follow-up papers have been built.
One such paper is \cite{kwon2020pomo} introducing POMO (Policy Optimization with Multiple Optima). There, the authors use the fact that, e.g., TSP has several solutions when using the different cities as start nodes.
Their approach helps with training and inference to obtain better solutions.
A further study extending the work of \cite{kool2018attention} is \cite{zhang2020deep}, who generalize to TSPTWR.
In recent work, \cite{jin2023pointerformer} propose to use reversible residual networks (\cite{gomez2017reversible}, \cite{kitaev2019reformer}) for their attention-based encoder to save memory.
By this, they can successfully process TSP instances of size up to 500 nodes. 
A similar idea is followed by \cite{yang2023memory} who propose a custom version of transformer models that is less memory extensive and can be applied to TSP with up to 1000 nodes.
Using earlier works of \cite{kool2018attention, kwon2020pomo, jin2023pointerformer} as a basis, \cite{lischka2024} shows the importance of preprocessing TSP instances by sparsification (i.e., deleting edges in the TSP graph) when using GNN or transformer encoders. Furthermore, they propose an ensemble-based encoder of different sparsification degrees which can improve overall performance.
Following up on \cite{kwon2020pomo}, \cite{kwon2021matrix} introduces a novel attention-based neural architecture suitable to operate on ``matrix'' structured combinatorial optimization problems like the distance matrices of TSP. By this, they can create a powerful network applicable to asymmetric TSP.
Also tackling asymmetric TSP, \cite{lischka2024greatarchitectureedgebasedgraph} developed an edge-based GNN variant called graph edge attention network (GREAT) that is suitable for routing problems. In contrast to other, node-based GNNs that operate on node coordinates as inputs, the developed edge-based GNN directly operates on the edge distances.

In another recent work, \cite{ling2023deep} propose to create an image representation of the TSP and apply a convolutional neural network (CNN), a model that has traditionally been used in image processing settings.
In contrast to other papers, \cite{jiang2022learning} aim to tackle the fact that most studies assume uniform data distribution in the unit square. 
Via distributionally robust optimization, they retrain the architectures of \cite{joshi2019efficient} (SL) and \cite{kwon2020pomo} (RL) on pairs or sets of data distributions and achieve better generalization performance than the original architectures.
A further adaption of existing architectures is performed in \cite{sultana2021learning}. There, the architectures of \cite{kool2018attention} and \cite{nazari2018reinforcement} are changed to include an additional entropy loss term in the RL loss function for more exploration during learning. 
By this, both architectures can improve their performance.

In contrast to other papers, \cite{xin2020multidecoder} propose a Multi-Decoder Attention Model (MDAM) to train several policies at once which increases overall performance. 
Another work tackling larger instances is \cite{ma2019combinatorial} who generalize their framework to large TSP instances (up to 1000 nodes) and constrained TSP versions (i.e. TSP with time windows) by using a hierarchical framework with hierarchical RL.
One of the few papers using SL in incremental methods is \cite{xing2020graph} who train a GNN to predict probabilities for the next node to visit in a TSP given a partial solution.
These probabilities are used in an MCTS.

To achieve higher generalization performance, \cite{ouyang2021} and \cite{ouyang2021general} propose several novel RL training techniques.
These RL techniques include local searches improving the rewards and baselines as well as incorporating curriculum learning.
Another work trying to tackle real-world data is \cite{sultana2022} who introduce a framework based on a CNN combined with an LSTM.
It utilizes randomly generated instances (easy instances) to solve various common TSP instances (complex TSP instances).
Following the idea and architecture of \cite{kool2018attention}, \cite{xu2021} introduce another, improved transformer-based RL model which leads to better performance. 
A further work based on \cite{kool2018attention} is propsed by \cite{bresson2021} who suggest a RL framework for TSP with beam search decoding. The context embedding while decoding captures more information about the current partial solution, leading to a performance increase. 
A model specialized for several data distribution at once was developed by \cite{jiang2024}. They train a model with a single encoder but several decoders (all \cite{kool2018attention} based). Training is done such that each decoder is suitable for a different data distribution. 
A work using self-improvement was proposed by \cite{pirnay2024selfimprovement}. They use a transformer-based model in a self-improvement setting, the following way: several solutions are sampled by the current model, and the best one is used in a SL setting. The updated model can then be used for sampling again. 

\cite{mo2023} introduce a pair-wise attention-based pointer neural network for predicting drivers' delivery routes using historical trajectory data. In contrast to the typical encoder-decoder architecture, they utilize a novel attention mechanism for local pair-wise information between stops. To enhance route efficiency, they developed an iterative sequence generation algorithm, applied after model training, to identify the most cost-effective starting point for a route.
A similar work to \cite{deudon2018learning} was proposed by \cite{mele2021reinforcement}. Their framework extends the training pipeline by including information about a TSP graph's MST. 
By this, they achieve an increase in the overall architecture performance.
In \cite{hu2021}, a bidirectional graph neural network is proposed for the arbitrary symmetric TSP. By this, more realistic settings like sparse TSP graphs can be tackled.
A further study for more realistic settings is \cite{zhou2023}.
This work tackles VRPs of different distributions and sizes.
The authors introduce a meta-learning framework where a general model is trained, serving as the initialization of task-specific models at inference time. At inference time, the pre-trained model is adapted to the new task by further training. 

Works dealing with multi-objective TSP include \cite{perera2023}, which present a graph pointer network-based framework. The model can identify solutions on larger instances while being trained on small ones. 
Another work dealing with multi-objective combinatorial optimization problems (like TSP) is \cite{lin2022pareto}. Their approach can approximate the entire Pareto set of problems. Similarly, \cite{rydin2025} also tackles multi-objective routing problems but extends the setting to multi-graph and asymmetric problems using the GREAT network as an encoder. 
A further study within multi-objective optimization is \cite{li2020deep}, who propose to learn different models for different subproblems. A subproblem is a weighted combination of the different objectives, and the models for ``close'' problems share the parameters of the initialization of the training.
In \cite{ruiz2023}, a multi-agent reinforcement learning framework for the Prize-Collecting Traveling Salesman Problem (PC-TSP) is proposed. 
Their algorithm is online, which means a new Q table is learned for each PC-TSP instance and afterward used to find a solution.

Recent ML systems tackling the TSP face scalability issues in real-world scenarios with numerous vertices. \cite{mele2021} revisits ML application by focusing on a specific task immune to common ML weaknesses. We task the ML system with confirming the inclusion of edges most likely to be optimal in a solution. Leveraging candidate lists as input, the ML distinguishes between optimal and non-optimal edges, offering a balanced approach between ML and optimization techniques. The resulting heuristic, trained on small instances, extends its efficacy to produce high-quality solutions for large problems. 

A new approach is chosen by \cite{drakulic2024bq} based on bisimulation quotienting (BQ) in Markov decision processes (MDPs). 
By this, the states in the incremental process correspond to problem instances, making generalization of the trained model to much bigger instances during inference possible.
Building upon this idea, \cite{drakulic2024goalgeneralistcombinatorialoptimization} develops a multi-task model suitable for a variety of combinatorial optimization problems once trained.
A similar idea is followed by \cite{zhouMVMoE2024}, who also develop a multi-task model for different routing problems. In their architecture, they employ mixture-of-experts layers to tackle several routing problems at once.

\subsubsection{One-shot methods}

An overview of different works using one-shot methods can be found in Table \ref{tableOneShotCS}. For used abbreviations, we refer back to Table \ref{abbreviatons}.

\begin{table}[width=\linewidth,cols=6,pos=h]
\caption{One-Shot methods in routing problems}\label{tableOneShotCS}
\begin{tabularx}{\textwidth}{%
  >{\hsize=1.2\hsize}X
  >{\hsize=0.5\hsize}X
  >{\hsize=1.1\hsize}X
  >{\hsize=0.7\hsize}X
  >{\hsize=1.3\hsize}X
  >{\hsize=1.7\hsize}X
  >{\hsize=0.5\hsize}X
  }
\toprule
 & Problems & ML formulation & Target & ML technique & Novelty & Code\\
\hline
\cite{joshi2019efficient} & TSP & SL  & Heatmaps & GCN + BS & First heatmap-based approach  & Yes\\
\cite{fu2021} & TSP & SL & Heatmaps  & GCN + MCTS & Generalization to large instances & Yes \\
\cite{kool2022deep} & TSP, CVRP & SL & Heatmaps & GCN + DP & Combining DP and heatmaps & Yes\\
\cite{qiu2022dimes} & TSP & RL & Heatmaps & GNN + MCTS; samp. & Meta-learning and generalization to large instances & Yes\\
\cite{hudson2021graph} & TSP & SL & Regret & GNN + GLS & Regret-based GLS & Yes\\
\cite{goh2022} & TSP & RL & Heatmaps & TF + greedy dec.; BS & RL for heatmaps (no labels) & Couldn't locate\\
\cite{gaile2022unsupervised} & TSP & USL & Heatmaps & GNN + greedy & USL for heatmaps, asymmetric TSP & Couldn't locate\\
\cite{min2023unsupervised} & TSP & USL & Heatmaps & GNN + GLS & USL for heatmaps with good performance & Yes\\
 \cite{ye2023deepaco} & TSP, CVRP & RL, & ``Heatmaps'' & GNN + ACO,NLS & Ant Colonization Optimization approach & Yes\\
 \cite{sun2023difusco} & TSP & SL & ``Heatmaps'' & GNN + greedy/ MCTS & Diffusion Models & Yes\\
 \cite{xin2021neurolkh} & TSP, CVRP &  SL/USL & $\beta, \pi$ scores & GNN + ``LKH''& Predict candidate set for LKH & Yes\\
\bottomrule
\end{tabularx}
\end{table}

One of the pioneering works within one-shot methods was done by\cite{joshi2019efficient}. They use a graph convolutional network, GCN, to predict edge probability heatmaps for the TSP.
The network is trained to output the heatmaps via SL, where the ground truths were adjacency matrices encoding optimal tours.
To transform the heatmaps into valid tours, a beam search (\cite{medress1977speech}) is performed. 
In a follow-up work, \cite{kool2022deep} use the GCN of \cite{joshi2019efficient} to predict heat-maps for the TSP and extend the idea to CVRP and the TSP with time windows. 
They construct valid solutions from the heatmaps by dynamic programming.
Also following the idea of heatmaps, \cite{fu2021} generalize the concept to large TSP instances (up to 10000 nodes) by using a sampling-and-merging approach.
They predict heatmaps using a GNN based on \cite{joshi2019efficient} for subgraphs and merge the heatmaps afterward to solve the overall large instance.

In contrast to other papers, \cite{gaile2022unsupervised} proposed a loss formulation for the TSP which does not require target labels to produce heatmap-like outputs, making the approach unsupervised. 
In a recent work, \cite{min2023unsupervised} introduce another USL framework for heatmap generation. 
They train a Scattering Attention GNN (SAG; \cite{min2022hybrid}) to output transition matrices which can be transformed into heatmaps and minimize traveled distance while finding a Hamiltonian cycle and achieve state-of-the-art performance.
Also overcoming the need for training labels, \cite{qiu2022dimes} focus on the maximum independent set (MIS) problem and the TSP. 
They train a GNN by RL to predict heatmap-like outputs which are later used for sampling, greedy decoding, and MCTS.
The work of \cite{hudson2021graph} is different as they do not learn heatmaps but regret values for the edges in the TSP instance.
The predicted regret values are used in a guided local search to obtain valid tours for the TSP.
\cite{goh2022} adapt the framework of \cite{kool2018attention} to predict heatmaps in a one-shot method instead of an incremental method.
They keep the RL formulation, which leads to positive edge labels no longer being required.
In contrast to other works, ant colonization is used in \cite{ye2023deepaco}. They propose a framework combining deep learning and ant colonization optimization. The neural architecture is used to predict heuristic measures indicating how promising it is to include solution components in the solution of the routing problem.
The work of \cite{sun2023difusco} brings diffusion models to routing problems. Their trained diffusion model produces an output that is used similarly to a heatmap to find solutions by greedy decoding or MCTS. A different direction is chosen in the work of \cite{xin2021neurolkh}. They train a GNN to predict edge scores (by SL) and node penalties (by USL) later used in the search framework of the LKH algorithm. 

\subsection{Improvement-based approaches}
We now move on to papers dealing with \textit{improvement-based approaches} where solutions to routing problems are iteratively improved over and over again.

\subsubsection{Heuristic methods}
We provide a summary of heuristic works in Table \ref{tableHeuristicCS}. For the used abbreviations, compare Table \ref{abbreviatons}.

\begin{table}[width=\linewidth,cols=5,pos=h]
\caption{Heuristic methods in routing problems}\label{tableHeuristicCS}
\begin{tabularx}{\textwidth}{%
  >{\hsize=1.2\hsize}X
  >{\hsize=0.4\hsize}X
  >{\hsize=0.8\hsize}X
  >{\hsize=1\hsize}X
  >{\hsize=1.6\hsize}X
  >{\hsize=0.45\hsize}X
  }
\toprule
 & Problems & ML formulation & ML technique & Novelty &Code \\
\hline
\cite{chen2019learning} & CVRP & RL (DQL, AC) & LSTM + MLP + PN & Overall (metaheuristic) FW & Yes \\
\cite{d2020learning} & TSP & RL (PG AC) & GCN + LSTM + PN & Learning 2-opt & Yes\\
\cite{lu2020learning} & CVRP & RL (PG with BL) & TF + MLP & Learning different moves, better than LKH & Yes\\
\cite{wu2021learning} & TSP, CVRP & RL (PG AC) & TF & Learning 2-opt & Yes \\
\cite{ma2021} & TSP, CVRP & RL (PPO AC) & Enhanced TF & Adjusted TFs for positional encodings & Yes \\
\cite{kalatzantonakis2023} & CVRP & RL & BL + VNS & BL inspired VNS & Couldn't locate \\
\cite{da2021} & CVRP, TSP & RL (PG AC)  & GCN + LSTM + PN & Extension to other VRPs & Yes \\
\cite{sui2021} & TSP & RL (PG AC) & GNN + LSTM + PN/FiLM Net & Extension to 3-opt & Couldn't locate  \\
\cite{karimi2021}  & TSP & RL (Q-Learning) & Q-Table + Iterated LS & Q-Learning for improvement and Perturbation & Couldn't locate\\
\cite{parasteh2022} & TSP & RL (PG AC) & TF & SWA for better generalization and less forgetfulness & Couldn't locate \\
\cite{Hottung2019NeuralLN} & CVRP & RL (PG AC) & TF based & Learning to repair perturbated solutions in a LNS & Yes \\
\cite{yang2022} & CVRP & RL (PG AC) & TF + GNN + MLP & Learning to repair in LNS, larger instances & Couldn't locate\\
\cite{ma2024} & TSP, CVRP & RL (PPO AC) & Enhanced TF 
 + RDS & Flexible k-opt, exploring infeasible regions & Yes \\
\cite{zheng2021combining} & TSP & RL (Q-learning) & Q-table & Use Q-table for LKH candidate set & Yes \\
\bottomrule
\end{tabularx}
\end{table}

An early work of heuristic methods was performed by \cite{chen2019learning} who propose a framework called NeuRewriter, which consists of a region and a rule-picking step and has been applied to several combinatorial optimization problems, among them the CVRP.
In the case of the CVRP, the region and rule-picking steps both correspond to selecting a node in the current solution and moving one node after the other, thus generating a new solution.

A similar idea was followed in \cite{d2020learning} who propose a framework that incorporates 2-opt moves for the TSP. Their framework selects two node indices which are then part of the 2-opt move. This selection can be generalized to $k$-opt with $k>2$.
\cite{da2021} adapts the proposed method of \cite{d2020learning} for the TSP to two extensions: the multiple TSP and VRP, achieving results on par with classical heuristics and learned methods.
In \cite{sui2021}, the idea of \cite{d2020learning} is further generalized from 2-opt to 3-opt. 
The moves are learned by first selecting edges to remove and then determining a way to reconnect the resulting segments.
\cite{ma2024} trains a network to learn general $k$-opt moves (not only $k=2,3$ like in \cite{d2020learning,da2021, sui2021}) with the help of their new Recurrent Dual-Stream (RDS) decoder. 
In \cite{wu2021learning} 2-opt moves are learned for the TSP and the CVRP.
In particular, the authors train a transformer model to output probabilities for pairs of nodes at once (in contrast to \cite{d2020learning} where the nodes are chosen iteratively). 
In a current solution, the positions of the nodes of the selected pair are then swapped and the order of the nodes between them is flipped. 
Following up on \cite{wu2021learning}, \cite{ma2021} present Dual-Aspect Collaborative Transformer (DACT), that learns node and positional embeddings separately to reduce potential noise and incompatible correlations when selecting improvement operators. By this, performance is improved compared to other transformer-based approaches like in \cite{wu2021learning}. 
The work of \cite{parasteh2022} is based on \cite{wu2021learning}. They adopt a stochastic weight averaging method to prevent agent forgetfulness during training and to achieve better generalization.

 In a pathbreaking study, \cite{lu2020learning} propose the ``learn to improve'' framework for the CVRP which was the first learning-based method to achieve better performance than the LKH3 algorithm.
Their RL-based framework consists of randomly selected perturbation operators (to avoid getting stuck in local minima) and learned improvement operators.
These learned improvement operators can be chosen by the attention-based improvement controller from a variety of improvement operators.

Different from other works, \cite{kalatzantonakis2023} present an approach where bandit learning is used to select improvement operators within a Variable Neighborhood Search.

Studies using Q-learning include \cite{karimi2021} who use Q-learning to find suitable improvement operators within an Iterated Local Search metaheuristic to solve the TSP.
A further such work is \cite{zheng2021combining} who use offline Q-learning to determine the candidate set later used in the LKH-algorithm. By this, speed and performance improvements are observed. 

\cite{Hottung2019NeuralLN} propose a large neighborhood search (LNS) for VRPs. 
An initial solution is updated by iteratively applying destroy and repair operators, where repairing is done by a neural framework.
The model gets as input the endpoints of incomplete (sub)tours that should be reconnected to form valid solutions when repairing, therefore, the model input size depends on the destruction degree (amount of tour endpoints that have to be reconnected) and not on the instance size.
\cite{yang2022} propose a similar framework as \cite{Hottung2019NeuralLN} and expand it to CVRP instances of sizes up to 2000.

\subsubsection{Subproblem-based methods}
The different subproblem-based studies are listed in Table \ref{tableSubproblemCS}. For the used abbreviations we refer back to Table \ref{abbreviatons} once more.

\begin{table}[width=\linewidth,cols=5,pos=h]
\caption{Subproblem-based methods in routing problems}\label{tableSubproblemCS}
\begin{tabularx}{\textwidth}{%
  >{\hsize=1\hsize}X
  >{\hsize=0.5\hsize}X
  >{\hsize=0.8\hsize}X
  >{\hsize=0.8\hsize}X
  >{\hsize=1.9\hsize}X
  >{\hsize=0.45\hsize}X
  }
\toprule
 & Problems & ML formulation & ML technique & Novelty & Code\\
\hline
\cite{cheng2023select} & TSP & RL (PG, GBL) & TF & Improving subpaths of large TSP & WIP\\
\cite{kim2021learning} & TSP, CVRP & RL (PG, GBL, entropy loss) & TF & Learning two collaborative Policies & Yes \\
\cite{zong2022rbg} & CVRP & RL (PG, BL) & MLP + LSTM & Learning to merge SR for optimization & Couldn't locate\\
\cite{li2021learning} & CVRP & SL & TF & Predicting improvement when optimizing SR & Yes\\
\cite{ye2024} & TSP, CVRP & RL (PG, BL) & TF and GNN & Learns both partitioning and subproblem solving & Yes \\
\cite{falkner2023bigfailenabling} & CVRP & SL + RL & GNN + TF & Hybrid of GNN to select subgraphs and TF to optimize them  & Yes \\
\cite{luo2024selfimprovedlearningscalableneural} & TSP, CVRP & Self-SL & TF-based & Self-improved learning for partial tours with instances up to 100k cities & Couldn't locate \\
\bottomrule
\end{tabularx}
\end{table}

A classical representative for subproblem-based methods is \cite{cheng2023select} where the authors try to solve large TSP instances.
In their framework, subpaths of a current solution are sampled.
These subpaths are then optimized by a deviation of the architecture of \cite{kool2018attention}, adapted to output the shortest path between the two given end-nodes of the original subpath while visiting all other nodes occurring in the original subpath.

A collaborative setting to solve subproblems was proposed in \cite{kim2021learning}.
The authors train two models at once (with RL): 
One model, which they call the seeder (model of \cite{kool2018attention}), is used to create a diverse set of initial solutions (corresponding to complete tours) with a loss function that incorporates an entropy reward.
The second model, which they call the reviser (adaptation of the model of \cite{kool2018attention}), optimizes sampled subpaths of the initial solution provided by the seeder.
The authors apply their architecture to the CVRP, the TSP, and the prize-collecting TSP.

In \cite{zong2022rbg} large CVRP instances are tackled.
They propose a learning-based routine to partition the overall problem into subproblems and solve these with either learning-based approaches or other heuristics. 
The learning part involves an LSTM to compute representations of subtours, which allows for merging regions with similar representations.
Partitions are split, solved, and merged several times leading to an iterative improvement of the solution.
In a ``homochronous'' work, \cite{li2021learning} also deal with large-scale CVRPs.
For each tour in a current solution, a centroid is computed.
These centroids are clustered with k-means.
Afterward, a transformer architecture (which was previously trained with SL) predicts a cost for each generated cluster.
The cluster that promises the biggest improvement compared to the current solution is optimized with the LKH3 algorithm (\cite{lkh3}).
A similar idea is proposed in \cite{falkner2023bigfailenabling} where they train a GNN to predict a ``potential'' for further optimization of a subgraph of a CVRP instance using SL. The subgraph is then optimized using the model of \cite{kwon2020pomo}.
In contrast, a study tackling several routing problems at once is \cite{ye2024} which applies to the TSP, CVRP, and PCTSP. They train two models: one to predict partitions of large graphs and the other one to improve subpaths in the problems.

\subsection{Exact-algorithm-based methods}

Either heuristics or ML only provide an approximate solution with no systematic ways to improve it or to prove optimality. In recent years, there has been a growing interest in integrating ML into exact algorithms on routing-related problems mostly for its benefits in speeding up solutions \citep{cappart2021,sun2021,morabit2021,morabit2023,wang2023}, by such as reduce the size of master problems/subproblems \citep{morabit2021,shen2022,morabit2023} and reducing the search space \citep{sun2021}. \cite{sun2021} tried to eliminate potential edges such that Concorde has a smaller search space and is faster. They did this by training an SVM to predict which edges are promising and which ones are not. \cite{cappart2021} using DRL and proximal policy optimization to learn an appropriate branching strategy. They proposed a general and hybrid approach, based on DRL and constraint programming, for solving combinatorial optimization problems, such as the TSP with time windows. In the existing learning-based column generation method, the ML model was integrated into a branch-and-price algorithm to reduce the size of master problems/subproblems \citep{morabit2021,shen2022,morabit2023}, to capture human behavior \citep{bayram2022}, and to select a set of promising matches that are likely to develop into near-optimal routes \citep{wang2023}. Specifically, \citet{morabit2021} introduced a column selection approach using a binary classification model to solve crew scheduling and vehicle routing problems with time windows. This approach reduces computational time by predicting whether generated columns should be included in solving the restricted master problem. In an extension, \citet{morabit2023} developed an arc selection method for the same problems, identifying arcs likely to contribute to an optimal solution using binary classification. Furthermore, \cite{wang2023} proposed a crowdsourced last-mile delivery framework, incorporating parcel allocation and crowd-courier routing within a two-tiered system where crowd-couriers handle the final delivery leg. They presented a data-driven column generation algorithm leveraging machine learning to effectively identify a subset of feasible and high-quality routes from the route-based set-partitioning formulation.

\section{Machine learning applications in routing problems} \label{section5}

This section will focus on papers that use ML for routing problems in more practical or applied situations. In other words, the TSP and VRP are not included here. Specifically, it will cover only recent routing studies that employed ML methods. We can categorize the routing papers on ML applications in two ways: one based on focus areas, and the other based on ML approaches. An overview of emerging VRP variants is provided, along with a selection of industrial applications drawn from news and patents.

\subsection{Divided by routing focus areas}

Firstly, we can categorize the routing papers on ML applications based on routing focus areas. The listed characteristics include Heterogeneity (Het.), EV, TW, PDP, Collaboration (Collab.), Time-dependency, Dynamics, Stochastic, and Integration. Refer to Table \ref{characteristics} for a detailed description of these characteristics.

\begin{table}[width=\linewidth,cols=6,pos=h]
\caption{Description of typical characteristics in routing problems}\label{characteristics}
\begin{tabular*}{\tblwidth}{@{} LL@{} }
\toprule
Characteristic & Detailed description \\
\hline
Het. & Considering vehicle heterogeneity (Het.), specifically differences in types or attributes\\
EV & Electric vehicles (EV) are utilized, typically charging is incorporated\\
TW & Customers-specific time windows (TW) are considered\\
PDP & Pickup and delivery (PDP), a routing subset, involves collecting and delivering items\\
Collab. & Collaborative (Collab.) routing coordinates multiple vehicles or agents to optimize delivery routes\\
Time-dependent & The routing network changes over time, necessitating real-time considerations and dynamic adjustments\\
Dynamic & Conditions in routing scenarios fluctuate over time, such as traffic flow, demand, or resource availability\\
Stochastic & It uses probabilistic models to handle uncertainty and make decisions accordingly\\
Integrated & Routing as one component in integrated problems, e.g., location-routing problem\\
\bottomrule
\end{tabular*}
\end{table}

The studies on routing problems that consider more practical situations using ML are summarized in Table \ref{tableapplications}, mainly focusing on the problem characteristics.

\begin{table}[width=\linewidth,cols=6,pos=htbp]
\caption{Characteristic of ML-based routing problems}\label{tableapplications}
\begin{tabular*}{\tblwidth}{@{} LLLLLLLLLL@{} }
\toprule
 & Het. & EV & TW & PDP & Collab. & Time-dependent & Dynamic & Stochastic & Integrated\\
\hline
\cite{joe2020} & & & & & & & \texttimes & \texttimes &\\ 
\cite{salama2020} & \texttimes & & & & \texttimes & & & & \texttimes \\
\cite{zhang2020multi} & & & \texttimes & & & & & &\\
\cite{aljohani2021}  & & \texttimes & & & & & & &\\
\cite{basso2021} & & \texttimes & & & & \texttimes & & & \\
\cite{furian2021} & & & \texttimes & & & & & \texttimes &\\
\cite{li2021} & \texttimes & & & & & & & & \\
\cite{li2021heterogeneous} & & & & \texttimes & & & &\\
\cite{lin2021}  & & \texttimes & \texttimes & & & & & &\\
\cite{ma2021hierarchical} & & & & \texttimes & & & \texttimes & &\\
\cite{morabit2021} & & & \texttimes & & & & & &\\
\cite{qin2021} & \texttimes & & & & & & & &\\
\cite{wang2021} & & & \texttimes & & & & \texttimes & &\\
\cite{wu2021reinforcement}  & \texttimes & & & & \texttimes & & & &\\
\cite{wu2021solving}  & & & \texttimes & & & \texttimes & & &\\
\cite{xiang2021} & & & & & & & \texttimes & &\\
\cite{zhang2021}  & & & & \texttimes & & & \texttimes & &\\
\cite{alcaraz2022} & & & & & & & \texttimes & \texttimes &\\
\cite{basso2022}  & & \texttimes & & & & & \texttimes & \texttimes &\\
\cite{bogyrbayeva2023}  & \texttimes & & & &\texttimes & & & &\\
\cite{chen2022}  & &\texttimes & \texttimes & & & & & &\\
\cite{chen2022deep} & \texttimes & & & & \texttimes & & \texttimes & & \\
\cite{liu2022} & \texttimes & & & & \texttimes & & & \texttimes &\\
\cite{lu2022} & \texttimes & & & & \texttimes & & & & \\ 
\cite{ma2022} & & & & \texttimes & & & & & \\
\cite{niu2022} & & & & & & & & \texttimes & \\
\cite{qi2022} & & & \texttimes & & & \texttimes & & &\\
\cite{wang2022} & & & \texttimes & \texttimes & \texttimes & & & & \\
\cite{zhang2022} & & & \texttimes & & & & & &\\
\cite{zhou2022} & & \texttimes & \texttimes & \texttimes & & & \texttimes & &\\
\cite{dieter2023} & & & \texttimes & & & & & & \\
\cite{florio2023} & & & & & & & & \texttimes &\\
\cite{guo2023}  & & & & & & \texttimes & & &\\
\cite{liu2023route}& & & \texttimes & & & & & & \texttimes\\
\cite{mak2023} & & & & & \texttimes & & & &\\
\cite{pan2023} & & & & & & & \texttimes & & \\
\cite{vansteenbergen2023} & \texttimes & & \texttimes & & \texttimes & & & \texttimes  &\\
\cite{wang2023} & & & & & & & & & \texttimes\\
\cite{zhang2023} & & & \texttimes & \texttimes & & & & &\\
\cite{feijen2024} & & & \texttimes & & & & & &\\
\cite{guo2024} & & & & & & & & & \texttimes\\
\cite{li2024} & & & \texttimes & & \texttimes & & & \texttimes & \texttimes\\
\cite{shelke2024} & & & \texttimes & & & & & & \texttimes\\
\cite{van2024} & & & \texttimes & & & & & & \texttimes\\
\cite{wu2024} & & & & \texttimes & & & \texttimes & & \\
\cite{wu2024neighborhood} & & & \texttimes & & & & & &\\
\cite{xiang2024} & & & & \texttimes & & & \texttimes & &\\
\bottomrule
\end{tabular*}
\end{table}

\subsection{Divided by ML approaches}

Based on the proposed classification of ML methods shown in Fig. \ref{FIG:illustration}, the routing papers on ML applications are summarized in Table \ref{tableML}. It is evident that, in general, there are more improvement-based studies than construction-based ones. Compared to one-shot methods, incremental methods are more prevalent. Among the improvement-based approaches, heuristic-based methods have become particularly popular in recent years.

\begin{table}[width=\linewidth,cols=3,pos=htbp]
\caption{Applied routing papers categorized by ML approaches}\label{tableML}
\begin{tabularx}{\textwidth}{%
  >{\hsize=0.5\hsize}X
  >{\hsize=0.6\hsize}X
  >{\hsize=1.9\hsize}X
  }
\toprule
Main category & Secondary category & Papers\\
\hline
\multirow{7}{*}{Construction-based} & \multirow{5}{*}{Incremental} & \cite{aljohani2021,li2021,li2021heterogeneous,lin2021,wu2021reinforcement,wu2021solving,zhang2021,chen2022,liu2022,bogyrbayeva2023,guo2023,liu2023route,pan2023,vansteenbergen2023,zhang2023,levin2024,wu2024,xiang2024}\\
& \multirow{2}{*}{One-shot} & \cite{zhang2020multi,basso2022,ottoni2022,zhou2022,mak2023}\\
\hline
\multirow{6}{*}{Improvement-based} & \multirow{5}{*}{Heuristic-based} & \cite{joe2020,salama2020,ma2021hierarchical,qin2021,wang2021,xiang2021,chen2022deep,lu2022,ma2022,niu2022,qi2022,dieter2023,guo2024,feijen2024,wu2024neighborhood,shelke2024,li2024}\\
& Subproblem-based & \cite{basso2021,wang2022,van2024}\\
\hline
\multirow{2}{*}{Exact-algorithm-based} & & \cite{morabit2021,furian2021,zhang2022,wang2023,florio2023}\\
\bottomrule
\end{tabularx}
\end{table}

\subsubsection{Construction-based approaches}

Tables \ref{tableConstruction} and \ref{tableConstruction2} provide a summary of ML approaches applied in routing problems, with a focus on construction-based methods. In evaluating these studies, we considered factors such as the problems they addressed, the ML methods used, the specific ML formulations, and the techniques applied.

\begin{table}[width=\linewidth,cols=6,pos=h]
\caption{Construction-based ML in routing problems}\label{tableConstruction}
\begin{tabularx}{\textwidth}{%
  >{\hsize=1.2\hsize}X
  >{\hsize=1\hsize}X
  >{\hsize=0.6\hsize}X
  >{\hsize=1.4\hsize}X
  >{\hsize=0.8\hsize}X
  >{\hsize=1\hsize}X
  }
\toprule
 & Problems & Subcategory & ML method & ML formulation & ML technique  \\
\hline
\cite{zhang2020multi} & VRPTW & One-shot & RL, Multi-Agent Attention Model  & RL (PG)  & TF-based\\
\cite{aljohani2021} & EVRP & Incremental & Double Deep Q-learning & RL (DDQL) & MLP \\
\cite{li2021} & Heterogeneous CVRP & Incremental & DRL & RL (PG) & Custom, TF-based\\
\cite{li2021heterogeneous} & PDP & Incremental & DRL & RL (PG) & TF-based\\
\cite{lin2021} & EVRPTW & Incremental & DRL & RL (PG) & GNN, Attention + LSTM\\
\cite{wu2021reinforcement} & Truck-and-drone & Incremental & Encoder–decoder framework with RL & RL (PG) & TF\\
\cite{wu2021solving} & Time-dependent TSPTW & Incremental & DRL & RL (PG) & RNN + attention\\
\cite{zhang2021} & DynamicTSP, Dynamic PDP & Incremental & DRL & RL (PG) & Custom TF-based\\
\cite{basso2022} & Dynamic stochastic EVRP & One-shot & Safe RL & RL (Q-Learning) & Q-Table\\
\cite{chen2022} & EVRPTW & Incremental & DRL & RL (PG) & GAT (a GNN type)\\
\cite{liu2022} & Stochastic truck-and-drone &  Incremental & DRL, DQN and A2C & RL (DQL and AC) & MLPs\\ 
\cite{ottoni2022} & TSP with refueling & One-shot & RL & RL (Q-learning) & Q-table\\
\cite{zhou2022} & Dynamic EVRPTW & One-shot & Spatio-temporal graph attention network with a value decomposition-based multi-agent RL & RL (PG) &  GAT + attention \\
\cite{bogyrbayeva2023} & Truck-and-drone & Incremental & DRL, Attention-based encoder-decoder & RL (PG) & TF + LSTM (with attention) \\
\bottomrule
\end{tabularx}
\end{table}

\begin{table}[width=\linewidth,cols=6,pos=htbp]
\caption{\textit{(Continued.)} Construction-based ML in routing problems}\label{tableConstruction2}
\begin{tabularx}{\textwidth}{%
  >{\hsize=1.2\hsize}X
  >{\hsize=1\hsize}X
  >{\hsize=0.6\hsize}X
  >{\hsize=1.4\hsize}X
  >{\hsize=0.8\hsize}X
  >{\hsize=1\hsize}X
  }
\toprule
 & Problems & Subcategory & ML method & ML formulation & ML technique  \\
\hline
\cite{guo2023} & Time-dependent VRP & Incremental & DRL, Deep attention models & RL (PG) & TF-based\\
\cite{liu2023route} & LRP & Incremental & DRL, Hybrid Q-Learning-Network-based Method & RL (HQM) & Q-value-matrix\\
\cite{mak2023} & CoVRP & One-shot & Deep multi-agent RL & RL (PG) & MLPs\\
\cite{pan2023} & Dynamic VRP & Incremental & DRL & RL (PG) & GNN + RNN + attention\\
\cite{vansteenbergen2023} & Stochastic truck-and-drone & Incremental & DRL, Value/policy function approximation & RL (VFA, PFA) & MLPs\\
\cite{zhang2023} & PDPTW & Incremental & DRL, attention mechanism and encoder-decoder & RL (PG) & GNN + TF\\
\cite{levin2024} & Multi-truck VRP with multi-leg & Incremental & Encoder-decoder attention model & RL (PG) & TF-based\\
\cite{wu2024} & Dynamic PDP & Incremental & Bayes' theorem-based sequential learning & RL (VFA; DP) & Value-table\\
\cite{xiang2024} & Dynamic multi-vehicle PDP with crowdshippers & Incremental & DRL, Attention model with centralized vehicle network & RL (PG) & TF-based\\
\bottomrule
\end{tabularx}
\end{table}

\subsubsection{Improvement-based approaches}

Previously categorized, improvement-based approaches encompass two subgroups: heuristic-based and subproblem-based methodologies. Tables \ref{tableImprovebased} and \ref{tableImprovebased1} offer a summary of ML approaches applied in routing problems, emphasizing improvement-based methods. Note that improvement-based methods utilize not only learning methods but also non-learning methods. In other words, at least two methods are incorporated into solving the routing problems. For example, the subproblem-based methods will have one subproblem tackled by ML, and the other subproblem mostly solved by commercial solvers. Similarly, heuristic-based methods combine ML with heuristics. Therefore, we will summarize the ML and non-ML methods used and how ML contributes to the solution. In evaluating these works, we considered factors such as the problems they addressed, the role of ML, the ML and non-ML methods used, the specific ML formulations, and the techniques applied.

\begin{table}[!htp]
\small
\caption{Improvement-based ML in routing problems}\label{tableImprovebased}
\centering
\begin{tabularx}{\textwidth}{%
  >{\hsize=1\hsize}X
  >{\hsize=1.2\hsize}X
  >{\hsize=1\hsize}X
  >{\hsize=1.2\hsize}X
  >{\hsize=1.2\hsize}X
  >{\hsize=1\hsize}X
  >{\hsize=0.6\hsize}X
  >{\hsize=0.8\hsize}X
  }
\toprule
 & Problems & Subcategory & ML help & ML method & Non-ML method & ML formulation & ML technique \\
\midrule
\cite{joe2020} & Dynamic VRP & Heuristic-based & Approximate value function & Neural networks-based Temporal-Difference learning  & Simulated Annealing & RL (VFA) & MLP\\
\cite{salama2020} & Clustering and routing, truck-and-drone & Heuristic-based & Accelerate solution time & Unsupervised learning & Heuristic & USL (clustering) & Clustering algorithm \\
\cite{basso2021} & EVRP with chance constraints & Subproblem-based & Predict energy consumption & Probabilistic Bayesian ML & MILP solver & SL (Bayesian ML) & Bayesian Regression based\\
\cite{ma2021hierarchical} & Dynamic PDP & Heuristic-based & RL framework & Hierarchical RL & Heuristic operators & RL (PG) & GNN\\
\cite{qin2021} & Heterogeneous VRP & Heuristic-based & Improve the effectiveness and extract hidden patterns & Distributed proximal policy optimization & Meta-heuristics & RL & Convolutional + MLP layers\\
\cite{wang2021} & Dynamic VRPTW & Heuristic-based & Combine strategies & Ensemble learning & Evolutionary algorithm & SL & Ensemble of basic models \\
\cite{xiang2021} & Dynamic VRP & Heuristic-based & Predict visiting order & Pairwise proximity learning & Ant colony algorithm & SL & RBF network within ACO\\
\cite{chen2022deep} & Same day delivery, vehicle-and-drone & Heuristic-based & Evaluate state and routing & Deep Q-learning & Assignment and routing heuristics & RL (DQL) & MLP\\
\cite{lu2022} & Truck-and-drone & Heuristic-based & Cluster task & Bisecting K-means & GA, SA, OR-Tools & USL (clustering) & Bisecting k-means\\
\cite{ma2022} & PDP & Heuristic-based & Synthesize features, perform removal and reinsertion & Transformer-based encoder-decoder & Neighborhood search & RL (PPO) & TF-based\\
\cite{niu2022} & Stochastic VRP & Heuristic-based & Generate and test hypotheses & Radial basis function network & Multi-objective evolutionary algorithm  & SL  & RBF network\\
\cite{qi2022} & Time-dependent green VRPTW & Heuristic-based & Guide heuristic transitions & Q-learning & NSGA-II, ALS & RL (Q-learning) & Q-table\\
\bottomrule
\end{tabularx}
\end{table}

\begin{table}[!htp]
\small
\caption{\textit{(Continued.)} Improvement-based ML in routing problems}\label{tableImprovebased1}
\centering
\begin{tabularx}{\textwidth}{%
  >{\hsize=1\hsize}X
  >{\hsize=1.2\hsize}X
  >{\hsize=1\hsize}X
  >{\hsize=1.2\hsize}X
  >{\hsize=1.2\hsize}X
  >{\hsize=1\hsize}X
  >{\hsize=0.6\hsize}X
  >{\hsize=0.8\hsize}X
  }
\toprule
 & Problems & Subcategory & ML help & ML method & Non-ML method & ML formulation & ML technique \\
\midrule
\cite{wang2022} & CoVRPTW & Subproblem-based & Customer clustering & Improved 3D k-means clustering & Genetic algorithm, particle swarm optimization & USL (clustering) & K-means based \\
\cite{dieter2023} & TSPTW with deviation & Heuristic-based & Predict driver behavior & Feedforward neural network & VNS & SL & MLP\\
\cite{feijen2024} & VRPTW & Heuristic-based & Predict potential & Supervised classification model& Large Neighborhood Search & SL & Random forest\\
\cite{guo2024} & Inventory routing problem & Heuristic-based & Framework & RL, offline/online/persistent learning  &  Adaptive heuristic & RL (Q-learning)  & Q-table \\
\cite{li2024} & VRPTW by drone with parcel consolidation & Heuristic-based & Improve particles' quality & RL & Particle swarm optimization, neighborhood search & RL (Q-learning) & Q-table\\
\cite{shelke2024} & Sourcing and routing & Heuristic-based & Assign dynamic customers& Deep Q-learning & Heuristic & USL (AE) + RL(DQN) & Graph Auto Encoder + MLP \\
\cite{van2024} & VRPTW feasibility check & Subproblem-based & Predict and support time slot decisions & Supervised ML, random forests, NN, gradient boosted trees & Routing Solver, ORTEC & SL & Random forests, neural networks, and gradient boosted trees\\
\cite{wu2024neighborhood} & VRPTW & Heuristic-based & Speed up convergence & Learning probability/ exemplars & Particle swarm optimization & RL & Comprehensive learning particle swarm optimization \\
\bottomrule
\end{tabularx}
\end{table}

\subsubsection{Exact-algorithm-based methods}

Table \ref{tableExact} reviews studies integrating machine learning with exact algorithms in routing problems, focusing on how ML enhances optimization techniques such as branch-and-price and column generation. It highlights the role of ML in tasks like variable selection, boosting algorithm performance, and improving prediction accuracy across various routing scenarios.

\begin{table}[!htp]
\small
\caption{Exact-algorithm-based ML in routing problems}\label{tableExact}
\centering
{
\begin{tabularx}{\textwidth}{%
  >{\hsize=1\hsize}X
  >{\hsize=1.2\hsize}X
  >{\hsize=1.2\hsize}X
  >{\hsize=1.2\hsize}X
  >{\hsize=1\hsize}X
  >{\hsize=0.6\hsize}X
  >{\hsize=0.8\hsize}X
  }
\toprule
 & Problems & ML help & ML method & Non-ML method & ML formulation & ML technique \\
\midrule
\cite{furian2021} & VRPTW  & Variable and node selection & Learning-based prediction & Branch-and-price & SL & MLP or random forest or logistic regression\\
\cite{morabit2021} & Scheduling, VRPTW  & Accelerate CG & ML & Column generation (CG) & SL & GNN\\
\cite{zhang2022} & VRPTW with two-dimensional packing & Boost CG mechanism & SL & Branch-and-price& SL& MLP\\
\cite{florio2023} & VRPSD under optimal restocking   & Handle correlated demands & Bayesian-based iterated learning & Branch-price-and-cut & SL & Bayesian learning\\
\cite{wang2023} & Route-based set-partitioning  & Predict travel time & ML & Branch-and-price & SL & XGBoost\\
\bottomrule
\end{tabularx}
}
\end{table}

\subsection{Emerging VRP variants}

In this subsection, we will introduce some emerging VRP variants or areas worthy of further study, including routing connected to the grid, routing in various modes of transportation, and integrated problems where routing plays a significant role alongside other components.

\subsubsection{Routing connected to grid}

With the rise of electric vehicles (EVs), the electric vehicle routing problem (EVRP) has received increasing attention over the past decade. The EVRP extends the traditional VRP by incorporating battery constraints, charging operations, and energy consumption. Early work by \citet{conrad2011} introduced recharging at customers' locations. \citet{schneider2014} integrated customer time windows and recharging at stations using a full recharge strategy. Later, \citet{bruglieri2015}, \citet{desaulniers2016}, and \citet{keskin2016} explored partial recharge strategies. For a comprehensive review, see \citet{kucukoglu2021}, which classifies EVRP studies based on objective functions, energy consumption calculations, constraints, and fleet types.
Note that some studies of the EVRP incorporate energy consumption estimation models to calculate energy usage more accurately. These models consider factors such as travel speed, acceleration, and vehicle load \citep{basso2021,heni2023}.

The interplay between the transportation network and the power grid has gradually garnered attention. This focus is mainly due to the mutual influence and dependence between the two systems, which is of great significance for improving the sustainability, efficiency, and resilience of cities.
Research into the integration of power networks with routing encompasses several crucial aspects. Firstly, with the advent of EVs, scholars have turned their attention to the EVRP \citep{conrad2011,schneider2014,keskin2016, zhou2024collaborative}. Secondly, there is a focus on developing precise energy consumption estimation models to accurately calculate EV energy usage \citep{basso2019,basso2021,heni2023}. This enables more effective path planning and charging strategies for EVs. Additionally, charging stations act as vital hubs linking the power grid and transportation network. Therefore, it is paramount to strategically plan the location and power supply capacity of these stations to meet the charging demands of distribution vehicles while maintaining the stable operation of the power grid. Despite scholarly attention to the location and routing of integrated charging stations \citep{zhang2019,yang2022integrated,hung2022}, considerations regarding power supply capacity have often been overlooked. Recently, researchers have shown increasing interest in integrating charging scheduling and routing \citep{kasani2021,chakraborty2021,liu2022collaborative}. This entails considering factors such as charging stations with stable power supply and balancing grid load, representing a significant shift in research focus over the past two years.

Considering the charging of electric vehicles from the grid, often referred to as grid-to-vehicle (G2V), it is crucial to also discuss vehicle-to-grid (V2G) technology. Initially, V2G research focused on leveraging electric vehicle batteries to balance grid loads and improve energy efficiency \citep{kempton2005}. Thereafter, with advancements in electric vehicle technology and the growing use of renewable energy sources, studies on V2G have garnered considerably more attention \citep{tan2016,das2020,zhang2020joint,qin2023}.

\subsubsection{Routing in different modes of transportation}

Routing problems are commonly linked to road transport but actually encompass diverse modes of transportation, such as air \citep{desaulniers1997,gopalan1998}, maritime \citep{ronen2002,fagerholt2010}, rail \citep{cordeau1998}, urban public \citep{silman1974}, and multi-modal \citep{moccia2011} transport.

In the realm of aviation transport, the challenges associated with routing have garnered considerable attention, especially with the emergence of drones. A review on drone applications in last-mile delivery can be found in \cite{garg2023}. The routing of drones has been explored as a variation of the TSP and VRP \citep{ermaugan2022}. For more literature on drone routing, refer to \cite{khoufi2019}. Introducing drones to collaborate with trucks for delivery tasks leads to the truck-drone routing problem (TDRP), which can be seen as a generalization of the classic VRP. For a comprehensive overview of drone-aided routing, see the reviews by \cite{macrina2020} and \cite{chung2020}.

In maritime transport, it is worth mentioning the ship routing problem, often referred to as the "ship routing and scheduling problem." This tactical distribution challenge involves a ship or fleet serving multiple ports to pick up and deliver goods. It extends the TSP and can be solved using VRP techniques. The literature in this area is vast; see \cite{christiansen2013,ksciuk2023} for a survey.
Another interesting problem is the ship weather routing problem, which involves optimizing the path of a single ship traveling from port A to port B, considering variable conditions like weather and waves. A comprehensive review can be found in \cite{zis2020}.
This problem differs significantly from the ship routing problem mentioned earlier.

ML techniques have been employed extensively in numerous studies on drone routing or drone-aided routing \citep{salama2020,wu2021reinforcement,arishi2022,chen2022deep,ermaugan2022,liu2022,bogyrbayeva2023,vansteenbergen2023} and route planning within maritime transport \citep{li2023,liu2023data}.

\subsubsection{Integrated problems: routing as one component}

Integrated problems typically entail at least two distinct decisions, with routing playing a critical role. On one hand, vehicle routing decisions, primarily operational in nature, may intersect with strategic or tactical decisions made over an extended planning horizon. These challenges encompass broader considerations such as facility location, fleet composition, and inventory and production management. They give rise to widely studied problems such as the location-routing problem \citep{prodhon2014} and the inventory-routing problem \citep{bertazzi2012}, as well as issues involving fleet composition and size with routing \citep{hoff2010}, and the production-routing problem \citep{adulyasak2015}. On the other hand, vehicle routing decisions are also intertwined with scheduling and loading challenges, leading to the routing and scheduling problem \citep{cisse2017} and routing problems with loading constraints \citep{pollaris2015}.

In recent years, ML techniques have been applied to integrated problems, particularly in areas such as location clustering and drone-based routing \citep{salama2020}, the location-routing problem for mobile parcel lockers \citep{liu2023route}, parcel allocation and crowd routing \citep{wang2023}, inventory-routing for bike-sharing systems \citep{guo2024}, routing with parcel consolidation \citep{li2024}, sourcing and routing \citep{shelke2024}, and routing with feasibility check \citep{van2024}.

\subsection{Industry applications}

ML is attracting attention in the logistics and transportation industry, with reports mentioning its use in routing under real-world constraints. For example, reports highlight that DHL employs AI-driven software to optimize last-mile delivery operations, dynamically sequencing routes while considering real-time traffic and delivery constraints \citep{dhl2023}. Similarly, Uber's DeepETA has been reported to leverage Transformer-based architectures to enhance ETA predictions, indirectly supporting dynamic routing and dispatching \citep{uber2022}. While these systems are still evolving, these reports illustrate the growing role of ML in industrial routing optimization, where adaptability and scalability are crucial for success.

Several industry patents have been filed, highlighting various ML-driven approaches for routing and logistics. The patent by Alipay \citep{zhang2020patent} introduces an innovative method for solving complex routing problems. The approach leverages a Siamese neural network to assess the similarity between routing problem instances and pre-solved cases stored in a database. By identifying the most similar cases, the system retrieves precomputed solutions and selectively applies optimization techniques tailored to the problem's constraints. This hybrid method reduces computational costs while maintaining solution quality, enabling scalable and efficient routing in real-world applications. The patent by State Farm \citep{Williams2022patent} introduces an innovative method for dynamically optimizing vehicle routing by leveraging AI and deep learning to address real-world constraints. This method integrates real-time data (e.g., traffic, weather, and task updates) and predictive analytics to continuously adjust service sequences. The system prioritizes tasks based on revenue maximization, time windows, and operational constraints, dynamically recalculating routes as new tasks or environmental changes arise. This adaptive, data-driven approach enhances the flexibility and scalability of vehicle routing. The patent by Doordash \citep{Han2024patent} focuses on optimizing delivery routes and task assignments in real time, relying primarily on traditional algorithms for routing optimization. Although machine learning is not directly applied to solving routing problems, it supports tasks such as ETA prediction by leveraging historical and real-time data to enhance system performance. This hybrid approach effectively combines the reliability of traditional optimization methods with the predictive capabilities of machine learning, resulting in improved logistics efficiency.

\section{Discussion} \label{section6}

In this section, we review and discuss the key insights gained from the various aspects of learning methods applied to routing problems. We start by reviewing the performance of different ML methods. Then, we discuss the challenges related to data preparation and generalization. We also explore the limitations of current methods and the importance of benchmarking and standardization. Finally, we offer recommendations for improving solution quality evaluation and establishing consistent metrics for future research.

\subsection{Performance overview - a selection}

In the following, we provide a performance overview of selected studies from the different approaches and methods on the TSP and CVRP in Table \ref{performance_overview}. In this overview, we focus on the most promising, well-performing studies of different branches in our taxonomy. While we try to provide a variety of different top-performing studies, we note that this selection is arbitrary and other studies are achieving (almost) equally good performance.
We state the performances (solution quality in terms of optimality gap and runtimes) as reported in the papers.
The gap indicating the solution quality of the ML approaches is computed with respect to the optimal solutions of the problems when dealing with the TSP. In the case of the CVRP, the gap is computed with respect to LKH3 (\cite{lkh3}). Solutions by LKH3 are not necessarily optimal, but they are of high quality. Using optimal solvers as a baseline for CVRP is typically not possible due to runtime constraints.
The solved instances in the experiments of ML papers are typically generated synthetically by sampling coordinates in the unit square. 
This setting is unrealistic, but it makes comparing the different approaches easy.
While not reported in the table, some papers, e.g., \cite{min2023unsupervised, fu2021, cheng2023select} also solved TSP instances with thousands of nodes with optimality gaps $<5\%$.
Furthermore, we note that several studies apply subproblem-based methods on large CVRP instances with several hundreds to thousands of nodes (e.g., \cite{li2021learning, ye2024, zong2022rbg}). Many of these papers are based on the LKH algorithm, e.g., by using it as a subsolver for partial problems. By this, they achieve better performance than plain LKH.
To summarize, we can see that different methods excel at different tasks. Incremental methods work well on `small' TSP and CVRP instances and are extremely fast as they do not require additional search procedures.
One-shot methods are well-suited for the TSP and can achieve very good optimality gaps in this task.
Heuristic methods excel on (small) CVRP instances, achieving better performance than the LKH3 algorithm.
And subproblem methods are extremely well suited for large instances, especially CVRP.
We further provide a small overview of the performance of two studies tackling asymmetric TSP in Table \ref{performance_overview_atsp}.
Since not many works have studied ATSP yet, we only include two incremental methods that achieve good performance on the task.
We note that there are some further studies like \cite{ye2024} where a subproblem-based method based on \cite{kwon2021matrix} is used to solve larger ATSP instances. Further, \cite{drakulic2024bq, drakulic2024goalgeneralistcombinatorialoptimization} both generalize to larger ATSP instances as well ($8.26\%$ and $2.37\%$ gap on instances of size 1000 respectively).

\begin{table}[width=\linewidth,cols=5,pos=h]
\caption{Performance overview of selected papers with top performance}\label{performance_overview}
{
\begin{tabularx}{\textwidth}{%
  >{\hsize=1.1\hsize}X
  >{\hsize=1.2\hsize}X
  >{\hsize=0.6\hsize}X
  >{\hsize=0.65\hsize}X
  >{\hsize=0.55\hsize}X
  >{\hsize=1.9\hsize}X
  }
\toprule
 & Method & Problem & Optimality Gap & Runtime & Comment\\
 \hline
Concorde & Solver & TSP100 & 0\% & 1h & Time for 10k instances by \cite{kwon2020pomo}\\
LKH3 & Traditional Heuristic & TSP100 & 0\% & 25min & Time for 10k instances by \cite{kwon2020pomo}\\
\cite{kwon2020pomo} & Incremental & TSP100 & 0.14\% & 1min & Time for 10k instances, transformer encoder + attention-based decoder\\
\cite{min2023unsupervised} & One-Shot & TSP100 & 0\% & 10min & Time for 10k instances, GNN encoder + search decoding\\
\cite{ma2024} & Heuristic Method & TSP100 & 0.33\%/0\% & 17min/7h & Time for 10k instances, enhanced transformer encoder + recurrent dual stream decoder\\
\cite{kim2021learning} & Subproblem & TSP100 & 0.54\% & 4.3s & Time for 1 instance, transformer encoder + attention-based decoder\\
 \hline
Concorde & Solver & TSP500 & 0\% & 37min & Time for 128 instances by \cite{min2023unsupervised}\\
LKH3 & Traditional Heuristic & TSP500 & 0\% & 11min & Time for 128 instances by \cite{min2023unsupervised}\\
\cite{jin2023pointerformer} & Incremental & TSP500 & 3.56\% & 1min & Time for 128 instances, transformer encoder + multi-pointer network decoder\\
\cite{min2023unsupervised} & One-Shot & TSP500 & 0.85\% & 3min & Time for 128 instances, GNN encoder + search decoding\\
\cite{cheng2023select} & Subproblem & TSP500 & 2.40\% & 15s & Time for 1 instance, transformer encoder + attention-based decoder\\
\hline
LKH3 & Traditional Heuristic & CVRP100 & 0\% & 12h & Time for 10k instances by \cite{kwon2020pomo}\\
\cite{kwon2020pomo} & Incremental & CVRP100 & 0.32\% & 2min & Time for 10k instances, transformer encoder + attention-based decoder\\
\cite{kool2022deep} & One-Shot & CVRP100 & 1.71\%/0.41\% & 60min/49h & Time for 10k instances, GNN encoder + dynamic programming decoding\\
\cite{lu2020learning} & Heuristic Method & CVRP100 & -0.5\% & 24min & Time for 10k instances; better than LKH3, transformer encoder + MLP decoder\\
\cite{kim2021learning} & Subproblem & CVRP100 & 2.11\% & 1.73s & Time for 1 instance, transformer encoder + attention-based decoder\\
\bottomrule
\end{tabularx}
}
\end{table}

\begin{table}[width=\linewidth,cols=5,pos=h]
\caption{Performance overview of selected papers with top performance continued}\label{performance_overview_atsp}
{
\begin{tabularx}{\textwidth}{%
  >{\hsize=1.2\hsize}X
  >{\hsize=1.2\hsize}X
  >{\hsize=0.6\hsize}X
  >{\hsize=0.9\hsize}X
  >{\hsize=0.6\hsize}X
  >{\hsize=1.5\hsize}X
  }
\toprule
 & Method & Problem & Optimality Gap & Runtime & Comment\\
 \hline
CPLEX & Solver & ATSP100 & 0\% & 5h & Time for 10k instances by \cite{kwon2021matrix}\\
LKH3 & Traditional Heuristic & ATSP100 & 0\% & 1min & Time for 10k instances by \cite{kwon2021matrix}\\
\cite{kwon2021matrix} & Incremental & ATSP100 & 3.24\%/0.93\%& 34s/1h & Time for 10k instances, matrix encoder, attention-based decoder\\
\cite{drakulic2024bq} & Incremental & ATSP100 & 1.27\%/0.96\% & 1min/19min & Time for 10k instances, GNN + transformer architecture, good generalization\\
\bottomrule
\end{tabularx}
}
\end{table}

Many practical routing problems typically involve fewer than 50 customers, as seen in studies like \cite{wu2021solving}, \cite{basso2022}, and \cite{liu2022}. However, an increasing number of studies have extended the problem size to 100-150 customers, where comparisons between ML, OR, and heuristic methods are more common. Notable studies in this range include \cite{zhang2020multi}, \cite{li2021}, \cite{li2021heterogeneous}, \cite{lin2021}, \cite{zhang2021}, \cite{chen2022}, and \cite{wu2024neighborhood}. Some studies, such as \cite{bogyrbayeva2023}, \cite{ma2022}, \cite{guo2023}, \cite{vansteenbergen2023}, and \cite{li2024}, primarily compare ML with heuristics, typically at the 100-node scale. As the problem scale increases to 200-400 customers, some studies, such as \cite{wu2021reinforcement}, \cite{xiang2024}, and \cite{xiang2021}, compare ML with heuristics, while \cite{wang2023} and \cite{zhang2023} compare ML with both OR methods and heuristics. Additionally, \cite{florio2023} compares ML with exact algorithms. Finally, \cite{feijen2024} studies an even larger dataset of 1,000 customers, continuing to explore the comparative effectiveness of ML approaches.

In general, OR methods are suited for smaller-scale problems due to their exact nature but struggle with larger problems due to high computational cost. Heuristic algorithms are faster than OR methods but may not guarantee optimal solutions, though they remain valuable for larger-scale problems. In contrast, ML approaches are typically faster and provide competitive solutions, especially when compared to heuristics. However, comparing ML with heuristics may not offer significant new insights, as heuristics often lack reliable benchmarks. While these comparisons demonstrate the competitive performance of ML in routing optimization, challenges persist in handling larger datasets and ensuring consistent performance across varying problem scales. Additionally, the training process of ML models is often complex and time-consuming, with results that may not always be predictable, making it difficult to assess their effectiveness in real-world scenarios without extensive experimentation.

\subsection{Data for learning to route}
As pointed out in the previous section, ML-based frameworks to tackle routing problems are typically trained on synthetic data.
In particular, the coordinates of routing problems are often sampled uniformly at random in the unit square. Each node in the problem has associated coordinates $(x,y)$ with $x,y \in (0,1)$ uniformly at random. The coordinates are then used to compute Euclidean distances between the nodes.
Unfortunately, this data distribution is unrealistic, as the coordinates in real-world settings are often clustered. 
While real-world datasets are available (compare, e.g., for TSP, TSPLib \cite{reinelt1991tsplib}), these are often too small for training ML-based frameworks or offer too few instances of a specific size (e.g., there might not be enough instances with exactly 100 nodes each for training).
To overcome this limitation, we propose a data generator for Euclidean coordinates in the unit square that are not simply distributed uniformly at random. 
On the contrary, data sampled from our generator is typically clustered and can also contain grid- or line-shaped components. We provide examples of coordinates sampled by our generator in Figure \ref{fig:generator}.
Our generator is publicly available.\textsuperscript{2}
\renewcommand{\thefootnote}{}%
\footnotetext{\textsuperscript{2} https://github.com/Learning-for-routing/Benchmark-Generator}%
\renewcommand{\thefootnote}{\arabic{footnote2}} 
It works by first sampling coordinates uniformly at random in the unit square and then applying ten consecutive mutation operators. The used operators were proposed in \cite{bossek2019evolving} and are called \textit{explosion}, \textit{implosion}, \textit{cluster}, \textit{expansion}, \textit{compression}, \textit{linear projection}, and \textit{grid}. For an overview of how the individual operators influence the coordinates, we refer the reader to \cite{bossek2019evolving}.

\begin{figure}[ht!]\centering
\begin{tabular}{ccc}
\includegraphics[width=0.31\textwidth, trim=4 4 4 4, clip]{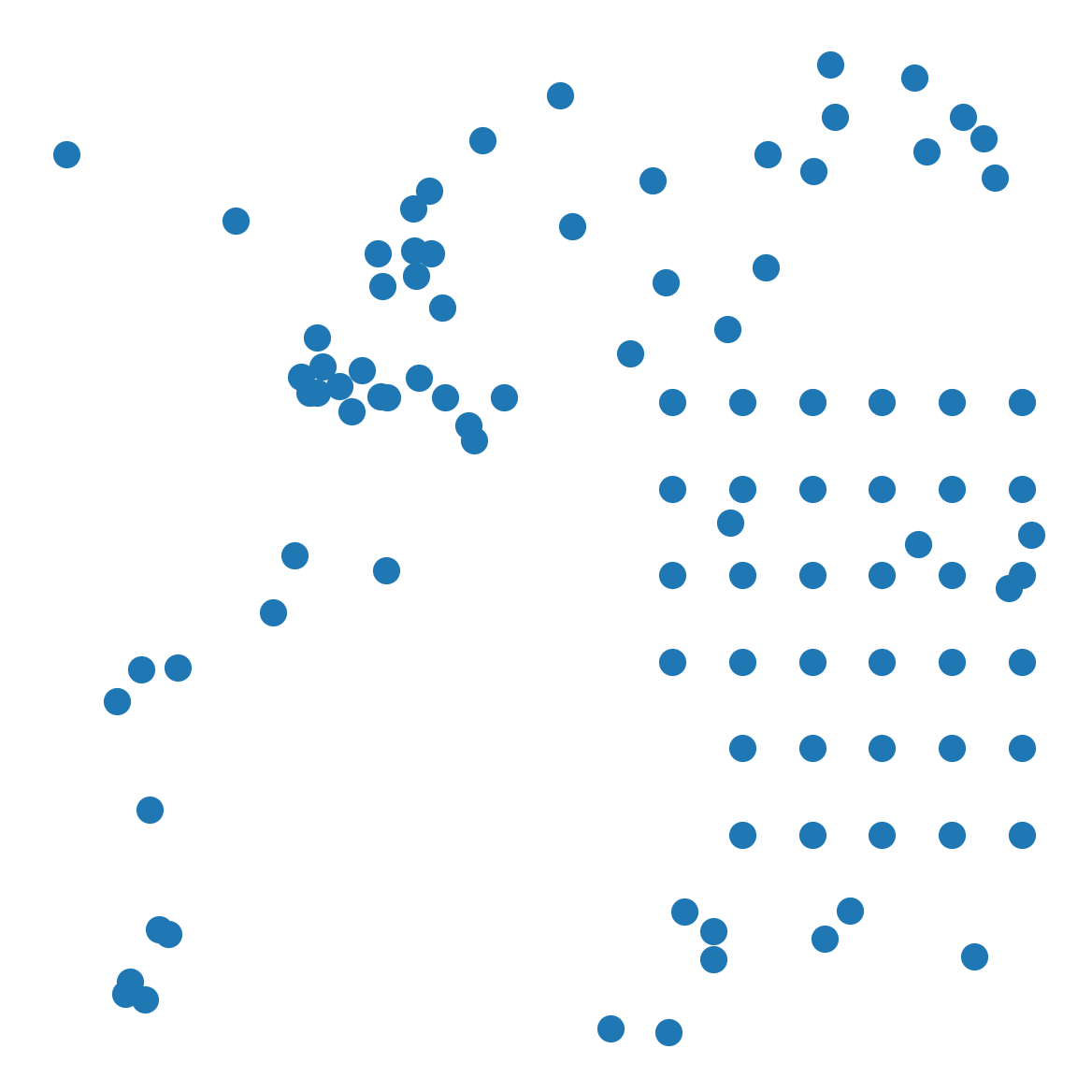} &
\includegraphics[width=0.31\textwidth, trim=4 4 4 4, clip]{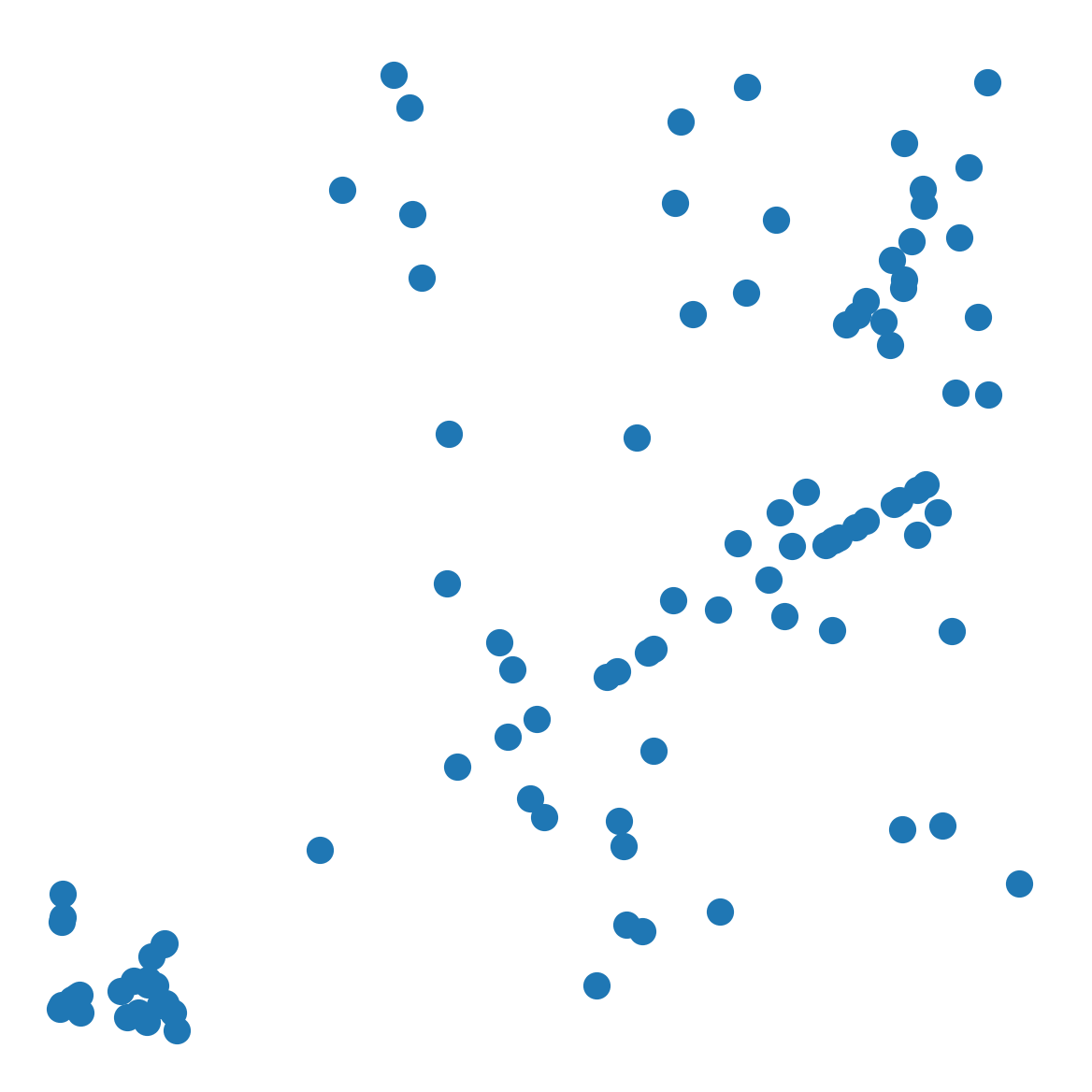} & \includegraphics[width=0.31\textwidth, trim=4 4 4 4, clip]{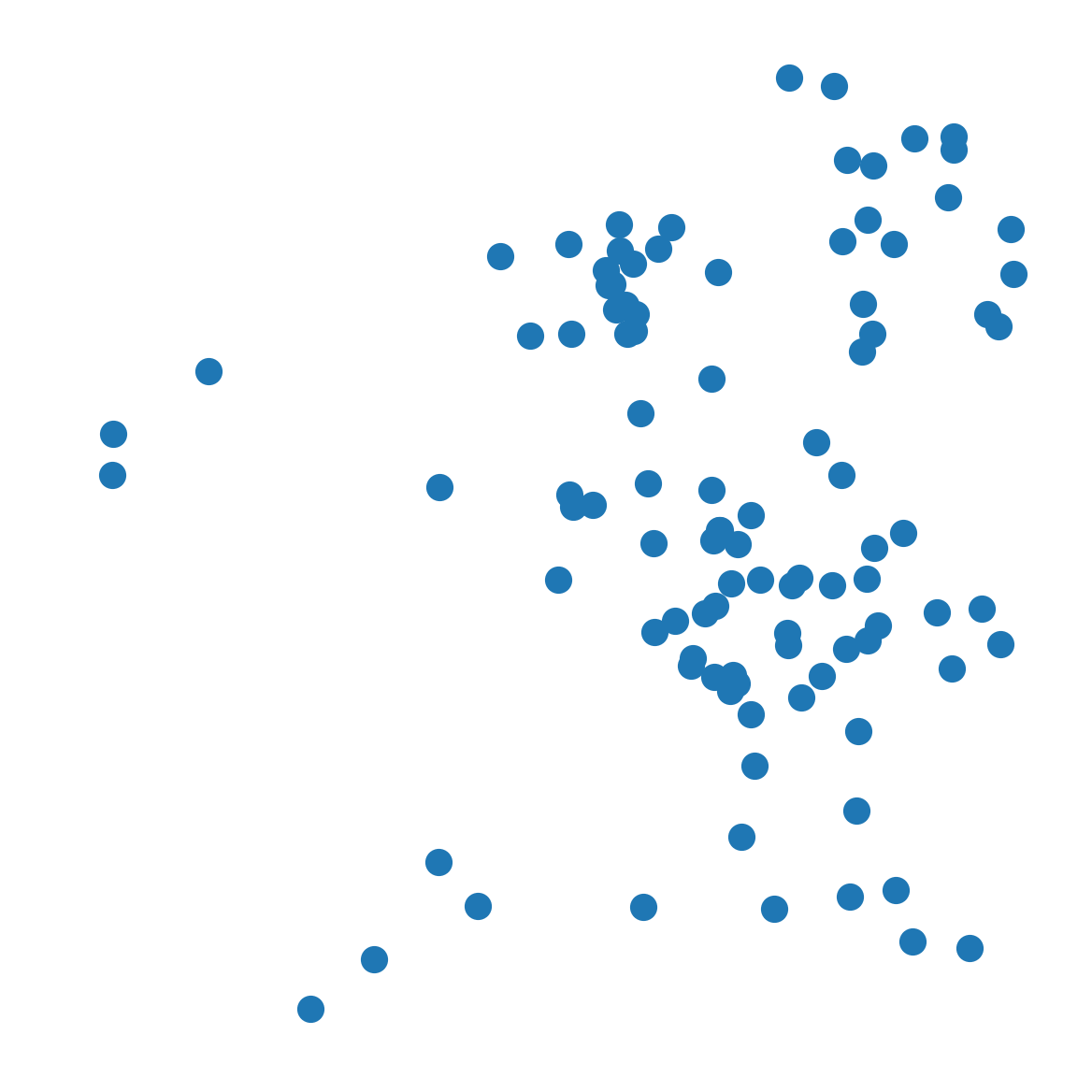} \\
(a) Example 1 & (b) Example 2 & (c) Example 3
\end{tabular}
\caption{Three data instances with 100 nodes each generated by our proposed data generator.}
\label{fig:generator}
\end{figure}

\subsection{Limitations and generalization challenges}

A key challenge in ML-based routing lies in addressing the exploration-exploitation trade-off. RL methods, such as deep Q-learning, often rely on epsilon-greedy strategies for exploration but may prematurely converge to suboptimal policies, particularly in sparse reward settings. On the other hand, in policy-based methods, the stochastic nature of the policy renders an inherent exploration. Other exploration methods could be entropy regularization (e.g., in Proximal Policy Optimization (PPO), Trust Region Policy Optimization (TRPO), and Soft Actor-Critic (SAC)), softmax policy, and exploration through noise. However, these exploration schemes usually require well-tuned settings that might generalize well to diverse datasets and environments. Similarly, neural architecture search (NAS) faces difficulties in balancing the discovery of new architectures and the optimization of known high-performing ones. Generative models, like GANs, may overly exploit existing modes of data distribution, leading to mode collapse. Advanced strategies, such as uncertainty-aware exploration, dynamic exploration schedules, and diversity-promoting objectives, have shown promise in mitigating these challenges and enhancing the generalization and robustness of learning-based approaches.

Furthermore, the reliance on uniformly distributed synthetic datasets significantly limits the generalization capabilities of ML-based frameworks, particularly in real-world routing problems with clustered or irregularly distributed locations. Real-world datasets like TSPLib are often too limited in size and diversity to support robust training, forcing many studies to depend on oversimplified synthetic data. These datasets fail to capture critical complexities, such as spatial clustering and geographic constraints, leading to models that perform well on synthetic benchmarks but may struggle to generalize to practical applications. To address these limitations, we propose a data generator that produces realistic distributions by incorporating features like clusters and grid patterns, enhancing the representativeness of training data. Such data generators can help bridge the gap between synthetic and real-world scenarios, providing a more effective and accurate way to train ML models. Future work could integrate such generators with real-world datasets to further validate their effectiveness and develop standardized benchmarking frameworks that incorporate realistic data distributions for evaluating model performance across diverse scenarios.
In the future, we hope for similar ``realistic'' data generators to emerge for a variety of routing problems, possibly also focusing on non-Euclidean or asymmetric settings. Such settings have been of comparably little interest so far but have been studied in, e.g., \cite{kwon2021matrix, ye2024, drakulic2024bq, lischka2024greatarchitectureedgebasedgraph}.

Moreover, addressing cross-metric and cross-problem challenges will require future models to handle diverse data distributions and performance evaluations effectively. For example, when adapting models to new problem settings such as non-Euclidean or asymmetric problems, models must be evaluated not only for their accuracy but also for their scalability and adaptability across different problem types. Additionally, the transition from synthetic to real-world data often involves significant shifts in distribution, necessitating the development of more adaptable ML methods that can seamlessly generalize across various routing problems, including those not traditionally addressed by current datasets.

Other, persisting limitations and challenges include the development of models capable of solving routing problems of different sizes and tackling multiple different routing problems at once.
So far, solving problems of different sizes has been achieved by divide-and-conquer approaches that partition problems into subproblems of sizes familiar to the trained ML model, e.g., in \cite{fu2021} or \cite{ye2024}.
Further, some promising results have been made in terms of generalization between different routing problems or even different combinatorial optimization problems \cite{ drakulic2024goalgeneralistcombinatorialoptimization, zhouMVMoE2024}.

\subsection{Benchmarking and standardization}

A significant challenge in ML-based routing research is the lack of standardized benchmarking practices. Differences in data distributions, problem sizes, and constraints, coupled with inconsistent definitions of performance metrics, make it difficult to compare results across studies. This limitation becomes even more pronounced in large-scale experiments, where subtle differences in experimental setups can lead to conflicting conclusions about model performance. Therefore, in the following, we propose several guidelines to ensure meaningful and comparable benchmarking of developed ML-based routing problem solvers.

\subsubsection*{Solution quality evaluation}

Firstly, to ensure consistent and meaningful comparisons, we propose the use of two types of benchmark solutions:

\begin{itemize}
    \setlength{\itemsep}{0pt}
    \setlength{\parsep}{0pt}
    \setlength{\parskip}{0pt}
\item Optimal Solutions: For small-scale instances, globally optimal solutions obtained using mathematical optimization tools such as Gurobi or CPLEX can serve as a definitive baseline.
\item Community-Recognized Benchmark Solutions: For larger-scale instances, benchmark solutions derived from state-of-the-art heuristics or metaheuristics (e.g., LKH) can be used instead. These solutions must be publicly available, reproducible, and widely accepted.
\end{itemize}

When neither of these benchmarks is available, approximate solutions generated by the proposed model itself can serve as a reference. In such cases, it is essential to clearly define the evaluation metric. We propose the following:

\begin{equation}
\text{Optimality Gap (\%)} = \frac{\text{Model Solution} - \text{Baseline Solution}}{\text{Baseline Solution}} \times 100.
\end{equation}

\begin{equation}
\text{Relative Deviation (\%)} = \frac{\text{Model Solution} - \text{Approximate Solution}}{\text{Approximate Solution}} \times 100.
\end{equation}

We emphasize that approximate solutions should only be used as a practical compromise, and their results should be interpreted cautiously.

\subsubsection*{Computational efficiency and complexity metrics}

Secondly, to ensure fair and reproducible evaluations of computational efficiency, we propose the following practices:

1) Normalized Runtime: Instead of absolute runtime, normalized runtime should be used to benchmark algorithms relative to a widely recognized reference, such as LKH:

\begin{equation}
\text{Normalized Runtime} = \frac{\text{Runtime of Algorithm A}}{\text{Runtime of Reference Algorithm}}.
\end{equation}

2) Scalability Factor: To evaluate algorithm performance as the problem size increases, the scalability factor can be used:

\begin{equation}
\text{Scalability Factor} = \frac{\text{Runtime for Larger Problem Size}}{\text{Runtime for Smaller Baseline Size}}.
\end{equation}

For example, testing with 50, 100, 200, and 500 nodes allows the evaluation of runtime growth trends. Scalability factors close to size ratios (e.g., 2×, 4×) suggest suboptimal scaling, while lower values indicate better scalability.

3) Standardized Experimental Settings: To make the results of experiments comparable, a standardized experimental setting is needed: 

\begin{itemize}
    \setlength{\itemsep}{0pt}
    \setlength{\parsep}{0pt}
    \setlength{\parskip}{0pt}
\item Problem Size: Consistent problem sizes (e.g., 50, 100, 200, 500 nodes) should be used.
\item Constraints: Time windows, vehicle capacity, or other constraints should be clearly specified.
\item Hardware Environment: Hardware specifications (e.g., CPU/GPU model) and software frameworks need to be reported.
\end{itemize}

4) Model Complexity: To measure the complexity of a developed method, the following characteristics should be reported:

\begin{itemize}
    \setlength{\itemsep}{0pt}
    \setlength{\parsep}{0pt}
    \setlength{\parskip}{0pt}
\item Number of Parameters: This corresponds to the total number of learnable parameters in a model. For instance, DNNs have more parameters compared to simpler models such as decision trees, influencing both training time and memory usage.
\item Memory Usage: It should be measured how much memory is required to store the model and its intermediate results, which is crucial for deploying large models on resource-constrained devices or environments.
\item Training Resources: The exact training hardware that was used needs to be reported. Further, the amount of data and time that was needed for training needs to be specified.
\item Inference Time: While training time is important, inference time—how quickly the model can make predictions—is also a key factor, especially for real-time routing tasks.
\end{itemize}

\subsubsection*{Benchmark selection}

To facilitate benchmarking, we recommend selecting widely recognized datasets and algorithms tailored to the size and complexity of the problem. For datasets, TSPLib, CVRPLib, and Solomon instances offer a robust foundation for evaluating algorithms across diverse problem scales and constraints. For algorithms, exact solvers like Gurobi or CPLEX are ideal for small-scale problems, providing globally optimal solutions as definitive baselines. For medium-scale problems, specialized heuristics such as LKH or Google OR-Tools are efficient and reliable, making them widely accepted benchmarks. For large-scale or highly complex problems, learning-driven models like the Pointer Network and Attention Model are valuable for assessing scalability and generalization. This tiered approach aligns benchmarks with problem characteristics, enabling meaningful and consistent comparisons across studies.

Establishing standardized benchmarking practices is essential for advancing ML-based routing research. By incorporating consistent performance metrics, such as optimality gap, relative deviation, normalized runtime, and scalability factor, along with standardized experimental settings, researchers can ensure fair and meaningful comparisons across studies. These practices not only improve the reproducibility and transparency of research but also provide insights into the inherent efficiency and scalability of algorithms. Future work should focus on refining these standards and validating their applicability across a broader range of routing problems, enabling the research community to collaboratively advance the state of the art.

\section{Proposed research agenda} \label{section7}

In order to guide researchers, this section aims at catalyzing the clustered and presented research results in the form of a \emph{proposed research agenda}. As such, we first define global goals for scientists and practitioners working within the topics covered. Second, we intend to list goal-oriented research areas via priority-based research gap analysis. Note that this priority is subjective as it reflects the authors' opinions. Furthermore, assigned to each priority area, research problems with promising solution methods are proposed.

UN Sustainable Development Goals (SDGs) have been selected to orient the readers of this paper. The contribution of the union of the fields of ML and OR can be related implicitly to nearly all SDGs. We, however, suggest to cover SDGs 7, 9-13, and 17 mostly because of their explicit research area couplings. Additionally, SDGs 7, 9-13, and 17 are further regrouped into social (10, 17), ecological (13), and economic (7, 9, 11, 12) subclusters. These selected goals and the subclusters proposed are subjective.

As previously discussed, economic research aspects currently dominate the field. However, the authors of this survey recognize the growing need to steer research efforts toward societal and ecological goals. With this in mind, the following thematic areas are highlighted in this agenda.

\subsection*{Resilience and robustness (economical SDGs 7, 9, 11, 12)}

One of the future directions is resilience and robustness in methodology, which involves creating adaptive systems to handle environmental changes, enhancing network and supply chain resilience, and integrating AI and machine learning for risk prediction and response. 
As such, focus on data-driven stochastic, robust, and distributionally robust optimization techniques to maintain performance under uncertainty have to be increased \citep{rahimian2019}. This would contribute to cross-disciplinary research to ensure system stability and reliability across various applications \citep{wu2020emergency}. As such, predictive-adaptive, e.g., reoptimization-based learning techniques are promising ways. 

The trade-off between conservatism and computational complexity may be approached via deep learning methods \citep{bertsimas2019,jiang2022learning}.
From a more computational angle, developing efficient and robust distributed algorithms is crucial in routing to ensure real-time adaptability and computational efficiency even in large-scale network environments. We hint at proper decomposition algorithms to use \citep{tian2024} or to investigate hybrid or quantum technology-based solutions \citep{abbas2024}.

\subsection*{Cross domain methods and implementations (ecological goals connected to SDG 13, economical goals connected to SDGs 7, 9, 11, 12)}

So far, ML has been used to tackle basic routing problems like the TSP. Despite achieving good to almost perfect performance in small instances (with up to 100 nodes) with optimality gaps close to 0\% and some work generalizing to instances with up to thousands of nodes, the settings are typically very artificial and have little relevance for real-world problems. For example, most papers assume uniform distributions of customers/cities/nodes in the unit square. Only a few papers generalize to other distributions as well, e.g., \cite{hu2021, jiang2022learning, alcaraz2022}. Furthermore, almost all papers focus on Euclidean distances. Non-Euclidean, potentially asymmetric distances have not been at the forefront of the investigation yet. Note that such settings are highly relevant in real-world implementations, for example, energy consumption (an objective of high relevance in electric vehicle routing) is typically asymmetric. E.g., if city A is located at a higher elevation than city B, traveling from A to B costs less energy than the other way around. Similarly, distances between customers for a delivery driver in a city might be asymmetric, as one-way streets may result in different distances depending on the direction one is traveling. We believe overcoming these limitations presents an interesting challenge, with the potential to develop a 'general-purpose' routing solver for problems like the TSP or CVRP.

Multi-objective OR problems, either via traditional model-based or via ML techniques, gain grounds \citep{li2020deep, lin2022pareto,perera2023}. As optimizers intend to select the best solution that has to satisfy multiple and often conflicting interests (e.g., minimize the total emission and the customer's wait time with home delivery at the same time). A promising way is to learn knowledge during the evolution process \citep{niu2022}. The minimization (or synchronization) of competing mobility objectives calls for horizontal collaboration, e.g., via game theory or other collaboration logic \citep{zhou2025collaborative}.

A particularly promising application of horizontal collaboration lies in collaborative routing. This includes coordination across companies, such as through shared infrastructure or joint optimization \citep{mak2023,zhou2024collaborative}, and coordination within a company across multiple transport modes, for example, truck–drone systems \citep{liu2022,bogyrbayeva2023}. These approaches can reduce cost and energy consumption while improving system-level efficiency, yet they also introduce new challenges in synchronization, task allocation, and uncertainty handling.

With EVs, range-related learning and optimization challenges arise. Given the current state of battery technology, frequent charging remains a significant challenge.
Furthermore, the need for charging couples mobility to the energy sector, and as such, future research has to be devoted to the ramifications of large-scale electric fleets impacting the stability of the power grid \citep{hussain2021,li2023electric}. This entails developing predictive models and mitigation strategies to address potential challenges \citep{panossian2022}. Additionally, there is an urgent need to explore the optimal deployment and management of V2G-enabled EVs. By balancing grid loads \citep{li2021vehicle}, enhancing energy efficiency, and offering economic benefits to EV owners \citep{bae2024}, such strategies can contribute to grid stability and sustainability. Moreover, expanding the scope beyond EVs to encompass drones, electric rail vehicles, electric ships, and other electric transportation modes is essential for comprehensive grid integration and sustainable transport solutions \citep{perumal2022,zhao2022}.

Integrating renewable energy stands as a pivotal endeavor. The future development direction entails establishing solar- and wind-powered charging infrastructure to harness clean energy sources for transportation needs. Transitioning public transport to electric and hydrogen fuel cells further advances sustainability goals. Moreover, powering drones and ships with renewable energy underscores the broad applicability of clean energy solutions. Implementing smart grid solutions enhances charging efficiency, while microgrid and community solar projects support local transportation hubs, fostering self-sufficiency and resilience in energy supply. This holistic approach ensures that renewable energy integration serves as a cornerstone for sustainable transportation development.

\subsection*{Climate neutral solutions (ecological goals connected to SDG 13)}

The primary focus of future work is to seamlessly integrate decarbonization strategies into transportation. This involves optimizing logistics operations to facilitate the widespread adoption of EVs and other low-emission transport modes across rail, road, water, air, and pipeline networks. These low-emission alternatives encompass various options, such as hydrogen fuel cell vehicles, sail-assisted ships, and biofuel aircraft \citep{ren2024,wang2022integrated,hou2022}.
In the realm of EV adoption, the integration of electric mobility solutions into routing problems has proven effective in promoting sustainability. The development of sophisticated energy consumption models for various types of EVs across diverse transportation modes is imperative for incentive implementation. These models should incorporate real-time data and machine learning to adapt to dynamic variables like traffic and weather conditions. To better support the utilization of electric mobility, the establishment of smart charging infrastructure is necessary. Designing smart charging infrastructure that dynamically adjusts based on grid load, vehicle priority, and real-time energy prices using AI and IoT technologies will further enhance the transition to electric transportation \citep{qaisar2022}.

\subsection*{Human- and environment-centered methods (ecological goals connected to SDG 13, societal goals connected to SDGs 10,17)}

Furthermore, exploring the use of real-time data from IoT devices and sensors can enhance responsiveness to changing conditions, optimizing logistics operations through dynamic real-time integration, aiming for more efficient use of ecological resources. This approach allows for the dynamic adjustment of routes and other logistical decisions, thereby contributing to the efficient adoption of low-emission transport modes \citep{chen2021pragmatic}. In addition, investigating human-centered logistics solutions can further enrich logistics operations, considering the well-being and preferences of operators and customers. Incorporating human-centric approaches, such as ergonomic routing for drivers and personalized delivery schedules, can enhance user experience and efficiency in the transportation system, ultimately supporting sustainable transportation development \citep{sun2023human}.

Another very recent development can be attributed to the steep rise in the popularity of LLMs.
These models have been used for many language-related tasks (social aspects), but can also be used to solve certain logic-related tasks, like coding.
Despite such models often not being well suited for math-related tasks, they have also been tested for solving routing problems like TSP \citep{yang2024large, ye2024reevo, masoud2024exploringcombinatorialproblemsolving}.
The pipeline in \cite{yang2024large} and \cite{masoud2024exploringcombinatorialproblemsolving} the LLM is directly used to create solutions for TSP.
In contrast, \cite{ye2024reevo} develops LLM-based hyper heuristics.
LLM-based solvers for routing or other combinatorial optimization problems will probably rise further in popularity with the improvements of ``reasoning-based'' LLMs such as OpenAI o1 and o3 \citep{zhong2024evaluation}. Due to the differences among human-spoken languages, further research may show the generalizability of LLM across language trees. 

\subsection*{Social compliance: ethical and interpretable ML implementations (societal goals connected to SDGs 10,17)}

The integration of ML into routing optimization has demonstrated significant potential, exemplified by systems like Uber's DeepETA and DHL's AI-powered route planning. However, challenges remain, including adapting to dynamic and stochastic environments, such as unexpected traffic disruptions, time-sensitive deliveries, and changing constraints. Future research should prioritize the development of robust and generalizable ML models capable of integrating real-time data streams with adaptive learning techniques to enhance responsiveness and reliability. A promising direction lies in combining ML with classical optimization techniques, leveraging ML's predictive capabilities alongside the precision of optimization algorithms to effectively address multi-objective routing problems. This includes sustainability goals, such as reducing fuel consumption and emissions, to support greener and more efficient logistics networks. Furthermore, lightweight and efficient ML models, such as those designed for edge computing, could enable real-time routing decisions under resource constraints, making them more practical for industrial deployment. Lastly, exploring general-purpose models that can handle diverse routing problems, including multimodal transportation and dynamic delivery systems, represents another critical avenue for bridging the gap between research and scalable industrial applications. We would also like to note the importance of considering real-world constraints such as scalability, data sparsity, and real-time adaptability, as they are crucial for practical deployment in large-scale routing scenarios. The nature and significance of such constraints is heavily case-specific and at the heart of problem modeling. Unfortunately, these valuable details are often proprietary in industrial solutions, limiting their accessibility and making it challenging to bridge the gap between academic research and real-world applications.

\subsection*{Safety critical methods and implementations (economical goals connected to SDG 10)}

Future research should focus on integrating safety-critical considerations into ML-based routing methods, ensuring safe operation in dynamic, high-risk environments. In addition to model interpretability, which ensures transparency and trust in decision-making, it is crucial to address ethical concerns, such as privacy protection, fairness, and bias mitigation. Protecting user privacy through methods such as differential privacy and data anonymization will be essential to ensure that models handle sensitive data responsibly \citep{gadotti2024,bae2024}. By integrating these ethical considerations, ML models can be developed to meet societal standards, ensuring that decisions are both safe and ethically sound, particularly in high-stakes applications like autonomous driving, emergency response, and logistics \citep{alipour2023,jayasutha2024}. Moreover, robustness to uncertainty is essential, enabling ML models to make safe decisions under stochastic conditions, such as traffic disruptions or extreme weather. Safety verification techniques should be employed to validate that models consistently produce reliable and safe outcomes. Finally, ensuring adversarial robustness will protect against the potential exploitation of models, particularly in critical applications like autonomous vehicles and emergency response systems \citep{silva2020,perez2024}.

\section{Conclusion} \label{section8}

This article aims to collect, structure, and summarize the literature on utilizing ML methodologies in OR, with a particular focus on routing problems, presented in a user-friendly handbook style. We propose a comprehensive taxonomy for ML-based routing methods, dividing them into construction-based approaches (one-shot and incremental methods) and improvement-based approaches (heuristic and subproblem-based methods). Exact-algorithm-based methods are incorporated into both construction-based and improvement-based approaches. This classification aids in understanding the landscape and evaluating factors like ML formulation, techniques, novelty, and problem characteristics. The study also introduces emerging variants of the routing problems or areas worthy of further investigation.

We reviewed key insights on the performance of different ML methods and discussed the challenges related to data preparation and generalization. The limitations of current approaches were also explored, with an emphasis on the importance of standardized benchmarking and solution quality evaluation. Based on these findings, we provided actionable recommendations for improving evaluation metrics and ensuring consistency in assessing ML models for routing problems.

Finally, the paper proposes a potential research agenda aligned with global objectives, particularly the UN Sustainable Development Goals, to guide future research efforts. By providing insights into existing gaps and offering actionable recommendations, we hope this work can stimulate further research and innovation in ML-driven solutions for routing optimization, ultimately bridging the gap between academic research and real-world applications.

\section*{Acknowledgments}

This work was supported by the European Commission, Swedish Energy Agency, and VINNOVA through the project E-LaaS (F-ENUAC-2022-0003) and project ERGODIC (F-DUT-2022-0078). The partial support of the Transport Area of Advance is herewith acknowledged via the project COLLECT. Finally, the project LEAR: Robust LEArning methods for electric vehicle route selection sponsored by the Swedish Electromobility Center is also acknowledged.
The authors would like to thank the anonymous reviewers for their thoughtful and constructive comments on the draft. While some of the feedback presented challenges, it greatly contributed to enhancing the quality of the paper. The authors also thank Dr. Okan Arslan for his kind suggestions during the revision process.

\bibliographystyle{cas-model2-names}

\bibliography{reference}

\end{document}